\documentclass[letterpaper, 10 pt, conference]{ieeeconf}
\IEEEoverridecommandlockouts
\usepackage[vlined, ruled, boxed, linesnumbered]{algorithm2e}

\SetCommentSty{mycommfont}

\SetKwProg{Function}{}{}{}

\usepackage{graphicx}
\usepackage{graphics}
\usepackage{times}
\usepackage{amsmath}
\usepackage{amssymb}
\usepackage{float}
\usepackage{url}
\usepackage{subfigure}
\usepackage{caption}
\usepackage{multirow}
\usepackage{gensymb}
\usepackage{epstopdf}
\usepackage{wrapfig}
\usepackage{setspace} 
\usepackage[noend]{algpseudocode}
\usepackage{color}
\usepackage{cite}
\usepackage{booktabs}
\usepackage{stackengine}
\usepackage[font={small}]{caption}
\usepackage[colorlinks,
            linkcolor=blue,       
            anchorcolor=blue,   
            ]{hyperref}
\usepackage{tikz}
\usepackage{textcomp}
\usepackage{lipsum}
\usepackage{pifont}

\usepackage[utf8]{inputenc}

\newcommand{\cc}{\color[rgb]{0,0.6,0.3}\checkmark}
\newcommand{\xx}{\color[rgb]{0.6,0,0}{\ding{55}}}

\title{\LARGE \bf
Super Odometry: IMU-centric LiDAR-Visual-Inertial Estimator for Challenging Environments	
}

\author{Shibo Zhao$^{1}$, Hengrui Zhang$^{1}$, Peng Wang$^{2}$, Lucas Nogueira$^{1}$, Sebastian Scherer$^{1}$
	\thanks{$^{1}$ Shibo Zhao, Hengrui Zhang, Lucas Nogueira, Sebastian Scherer are with Robotics Institute, Carnegie Mellon University, USA, {\tt\scriptsize \{shiboz,hengruiz,lcasanov,basti\}@andrew.cmu.edu}}
	\thanks{$^{2}$ Peng Wang is with Faculty of Robot Science and Engineering, Northeastern University, China 
		{\tt\scriptsize \{pengwang\}@mail.neu.edu.cn}} 	
}

\begin{document}

\maketitle
\thispagestyle{empty}
\pagestyle{empty}


\begin{abstract}
We propose Super Odometry, a high-precision multi-modal sensor fusion framework, providing a simple but effective way to fuse multiple sensors such as LiDAR, camera, and IMU sensors and achieve robust state estimation in perceptually-degraded environments. Different from traditional sensor-fusion methods, Super Odometry employs an IMU-centric data processing pipeline, which combines the advantages of loosely coupled methods with tightly coupled methods and recovers motion in a coarse-to-fine manner. The proposed framework is composed of three parts: IMU odometry, Visual-inertial odometry, and LiDAR-inertial odometry. The Visual-inertial odometry and LiDAR-inertial odometry provide the pose prior to constrain the IMU bias and receive the motion prediction from IMU odometry. To ensure high performance in real-time, we apply a dynamic octree that only consumes 10\% of the running time compared with a static KD-tree. The proposed system was deployed on drones and ground robots, as part of Team Explorer's effort to the DARPA Subterranean Challenge where the team won $1^{st}$ and $2^{nd}$ place in the Tunnel and Urban Circuits \footnote{https://www.subtchallenge.com/results.html}, respectively.
\end{abstract}


\section{Introduction}
Multi-Modal sensor fusion is essential for autonomous robots to fulfill complex and dangerous missions such as perception in subterranean environments, industrial inspection and search and rescue. In these GPS-denied scenarios, darkness, airborne obscurants conditions (dust, fog and smoke), and lack of perceptual features are major challenges that currently hinder us from employing robotic systems for long-term autonomy.  To localize in such environments, LiDAR-based odometry\cite{loam,shan2020liosam,Zhao_2019} seems to be a suitable choice for robots since LiDAR sensors can provide high-fidelity 3D measurements. However, in structure-less environments such as long tunnels or the presence of obscurants (e.g. fog, dust, smoke), LiDAR-based approaches suffer to provide reliable motion estimation because of degeneracies and outliers. To handle these situations, integrating with additional sensors, cameras in particular, is also required. In spite of this, the use cases of the visual camera are limited to well illuminated environments. Therefore, we argue that both LiDAR-based \cite{loam,shan2020liosam,Zhao_2019}, visual-based \cite{VINS,zhao2020tptio}  or LiDAR-visual-based \cite{graeter2018limo,ebadi2020lamp} SLAM methods are not the optimal choices in challenging environments. 
\begin{figure}[ht!]
	\centering
	\includegraphics[width=1.0\columnwidth]{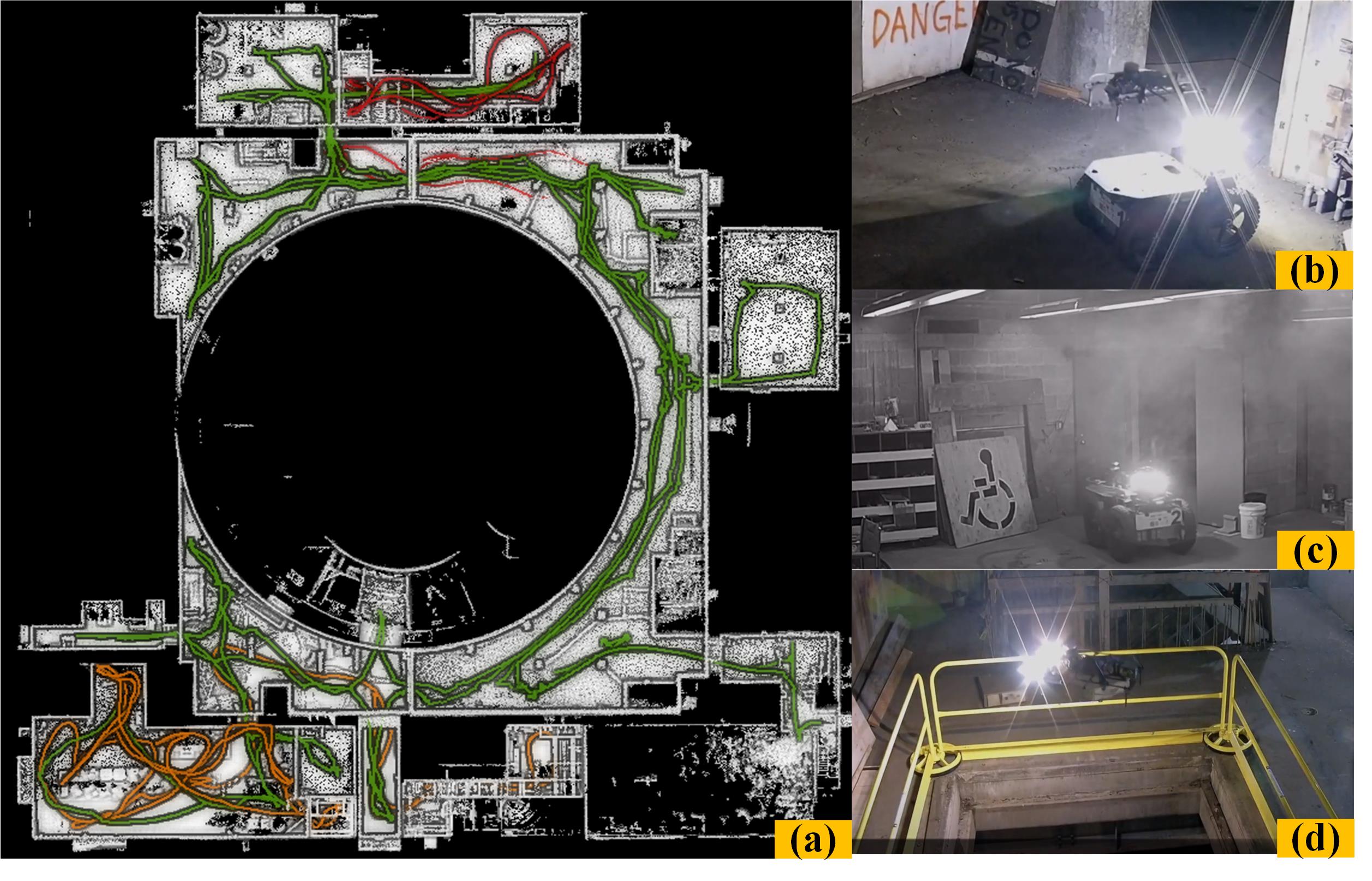}
	\caption{The Performance of Super Odometry in the DARPA Subterranean Challenge Alpha Course, which is challenging environments including darkness (b), heavy fog (c) and vertical shaft(d). (a) represents the mapping result of Super Odometry in Alpha Course reconstructed by multi robots (UGV1,UAV1 and UAV2); Green, orange and red lines are the estimated trajectory of UGV1, UAV1 and UAV2 respectively. 
	}
	\vspace{-3mm}
	\label{fig:alpha_course}
\end{figure}
Since these sensors are environment-dependent, it is difficult to achieve robust performance if they are used as dominant sensors to build a SLAM system.

In the last decade, multi-modal sensor fusion has gained great popularity and can be classified as either loosely or tightly coupled methods. As shown in Table \ref{tab::sensorfusion}, 
Loosely coupled methods \cite{loam,Zhao_2019,Palieri_2021} have been preferred more because of their simplicity, extendibility, and low computational expense. In contrast, tightly coupled methods \cite{shan2020liosam}  has proved advantageous for their accuracy and robust performance. However, tightly coupled methods are usually difficult to extend other sensors and achieve a switching scheme between sensing modalities to cope with various environments. In addition, tightly coupled methods may be vulnerable to cope with potential sensor failures since most of them only use a single estimation engine \cite{Palieri_2021}.
In comparison, loosely coupled methods distribute the sensor failure risks between several estimation engines and should have more robust performance. However, since most of the loosely coupled methods adopt frame-to-frame estimation, they still have limitations on accuracy and robustness.

Motivated by the discussion above, we proposed an IMU-centric SLAM system combing the advantages of tightly-coupled methods with loosely coupled methods. The system design follows a key insight: An inertial measurement unit (IMU) produces smooth measurements with noise but little outliers which can make its estimate very accurate as long as the bias drift can be well-constrained by other sensors.  Therefore, Super Odometry is designed around the IMU as the primary sensor. The main contributions of our work are as follows:

\begin{itemize}
	\item 	
	We propose the first IMU-centric sensor fusion pipeline that enables accurate real-time state estimation in extreme and perceptually-challenging environments.
	
	\item The proposed pipeline combines the advantages of tightly-coupled methods with loosely-coupled methods, which provides a simple but effective way to fuse multiple sensors.
	
	\item We propose to use an efficient organization of 3D points
         (dynamic octree) which significantly improves the real-time performance of scan matching.
	
	\item The proposed approach was deployed on multiple physical systems including drones and ground robots and has been extensive evaluated in various challenging scenarios with aggressive motion, low light, long corridor, and even in heavy dust environments.
\end{itemize}


\section {Related Work}
In recent years, there are a number of efforts have been made for the LiDAR-visual-inertial estimator, which can be classified as either loosely coupled methods or tightly coupled methods.

\begin{table}[]
	\centering
	\caption{The comparison of multi-modal sensor fusion algorithm }
	\label{tab::sensorfusion}
	\resizebox{0.95\columnwidth}{!}{
		\begin{tabular}{@{}ccccc p{0.2cm} @{}}
			\toprule
			Methods   & Accuracy  & Resilience  & Computation  &Extensibility  \\ \midrule
			Loosely coupled[1][2]  & \xx  & \xx  & \cc    & \cc        \\
			Tightly coupled [3][4][5][6]   & \cc & \cc & \xx    & \xx         \\
			Combination (Ours)  & \cc  & \cc   & \cc    & \cc         \\ \bottomrule
		\end{tabular}
	}
	\vspace{-3mm}
\end{table}

\subsection{Loosely Coupled LiDAR-Visual-Inertial Odometry}
Zhang and Singh \cite{zhang2018laser} proposed V-LOAM algorithm that adopts sequential data processing pipeline and used Visual-inertial odometry to provide motion prediction for LiDAR scan matching. However, since this pipeline still performed frame-to-frame motion estimation and the current estimate is totally based on the last estimate, it is difficult to achieve a recovery mechanism if the last estimate goes wrong. To solve this problem, Super Odometry adopts factor graph optimization and the current estimates are based on historical frames within a sliding window. Therefore, Super Odometry is fail-safe to single-point failures.
To improve the robustness, some works incorporate other constraints to LiDAR-visual-inertial estimator and obtain very promising results such as adding thermal-inertial prior\cite{multi_modal}, leg odometry prior \cite{camurri2020pronto} or loop closure \cite{Reijgwart_2020}. However, since these methods still performed frame-to-frame motion estimation, they have similar limitations with V-LOAM.

\subsection{Tightly Coupled LiDAR-Visual-Inertial Odometry}
To perform joint state optimization, Shao proposed VIL-SLAM \cite{shao2019stereo} which effectively fused Visual-inertial odometry and LiDAR scan matching within a graph optimization.
Zuo \cite{zuo2020licfusion} introduced LIC-Fusion which tightly coupled LiDAR edge features, sparse visual features, as well as plane features together by using the MSCKF framework\cite{msckf}.
David \cite{Wisth_2021} presents a unified multi-sensor odometry that jointly optimizes 3D primitives such as lines and planes. However, since the measurements of these methods are highly tightly coupled, they are difficult to extend to other sensors. In addition, since these methods usually use a single state estimation engine, they may have a higher failure rate when the sensors are damaged.
In contrast, Super Odometry is an IMU-centric sensor fusion pipeline and it only receives pose constraints from other odometry. Therefore, it is very easy to fuse other sensors. Also, since it has multiple state estimation engines, it has the capability to overcome potential sensor failure.

\subsection{Method Highlights}
In contrast with previous works, the advantages of Super Odometry are: 

\begin{itemize}
	\item \textbf{Robustness and Extendibility:}
	Our IMU-centric sensor fusion architecture (see Fig.2) enables us to achieve high accuracy and operate with a low failure rate, since the IMU sensor is environment-independent. The system includes fail-safe mechanisms and provides an easier and flexible way to fuse multiple sensors such as GPS, wheel odometry, and etc. 
	\item \textbf{Simple but Effective:}	
	 Super Odometry avoids complex derivations to fuse multiple sensors because the IMU has a simple and accurate probabilistic model. As long as other sensors can provide relative pose prior to constrain the IMU preintegration factor, these sensors will be fused into the system successfully. In particular, it has been extensively evaluated in challenging environments as part of Team Explorer's effort to the DARPA Subterranean Challenge.
	
	\item \textbf{Low CPU Usage and High Real-time Performance:} 
Since Super Odometry does not combine all sensor data into a full-blown factor graph, instead, it divided the big factor graph into several "sub-factor-graph" and each "sub-factor-graph" receives the prediction from an IMU pre-integration factor (see Fig.\ref{fig:algorithm}). Therefore, the motion from each odometry factor is recovered from coarse to fine manner in parallel, which significantly improves real-time performance. Also, Super Odometry adopts a dynamic octree to organize the 3D points, which makes scan-matching very efficient.

\end{itemize}


\section{System Overview}
\label{sec::system}

\begin{figure*}[htbp]
	\centering
	\includegraphics[width=1.85\columnwidth]{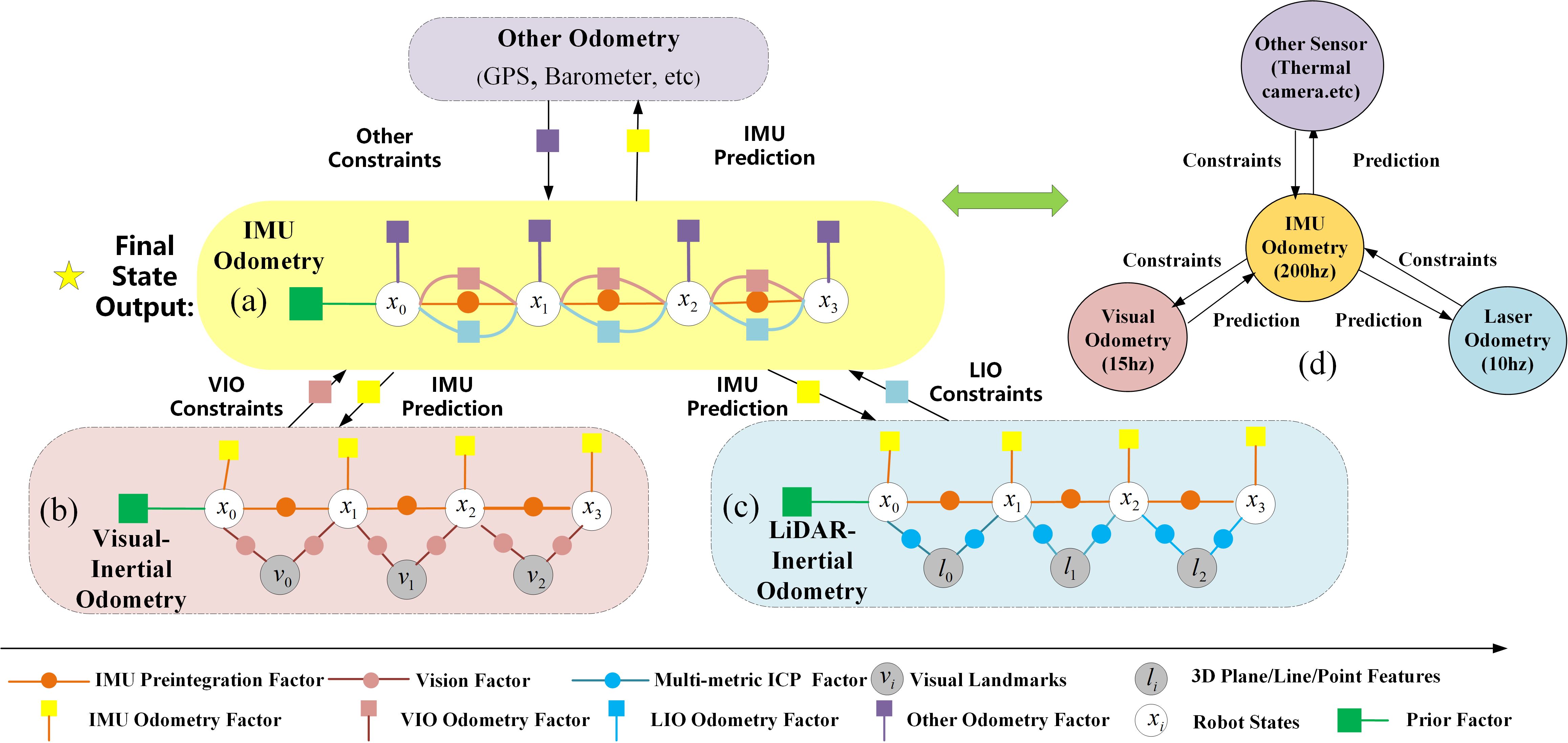}
	\caption{Overview of Super Odometry Alogorithm.
	IMU Odometry makes use of observations provided by visual odometry and LiDAR odometry to constrain the IMU bias. In return, the constrained IMU odometry provides the prediction to the visual odometry and LiDAR odometry. Meanwhile, the size of graph node can be adjusted according to the computational resource and the IMU Odometry provides the final state estimation.}
	\vspace{-3mm}
	\label{fig:algorithm}
\end{figure*}

We first define frames and notation that we use throughout the paper. We denote the world frame as  $\bold{W}$, IMU body frame as $\bold{B}$, LiDAR frame as $\bold{L}$ and camera frame as $\bold{C}$. For simplicity, we will regard the IMU frame as our robot body frame and our goal is to estimate the position of the IMU body frame relative to a fixed world frame $\bold{W}$.
The robot state $\mathbf{x}_{k}$  can be written as:

\begin{equation}
	\mathbf{x}_{k}=\left[\mathbf{p}_{b_{k}}^{w}, \mathbf{v}_{b_{k}}^{w}, \mathbf{q}_{b_{k}}^{w}, \mathbf{b}_{a}, \mathbf{b}_{g}\right], k \in[0, n]
\end{equation}

where the $k$-th robot state $\mathbf{x}_{k}$ consists of the position $\mathbf{p}_{b_{k}}^{w}$, velocity $\mathbf{v}_{b_{k}}^{w}$, orientation $\mathbf{q}_{b_{k}}^{w}$ of IMU's center, and $\mathbf{b}_{a}, \mathbf{b}_{g}$ are accelerometer bias and gyroscope bias respectively.


An overview of the proposed system is shown in Fig.\ref{fig:algorithm}. It is composed of three modules: IMU odometry, Visual-inertial odometry (VIO), 
and LiDAR-inertial odometry (LIO). Our system design follows a key insight: an IMU and its state estimation can be very accurate as long as the bias drift is well-constrained by other sensors. Therefore, Super Odometry uses the IMU as the primary sensor. It uses observations provided by Visual-inertial 
odometry (VIO) and LiDAR-inertial odometry (LIO) to constrain the accelerometer bias $\mathbf{b}_{a}$ and gyroscope bias $\mathbf{b}_{g}$. In return, the constrained IMU 
odometry provides the prediction to the VIO and LIO, which recovers motion in a coarse-to-fine manner. Moreover, the focus on the IMU makes the pipeline more robust to environment-related issues in state estimation, such as geometric or Visual degradation. Finally, the algorithm architecture allows for an easy and flexible way to incorporate other odometry sources such as GPS and wheel odometry.

\section{Methodology} \label{sec:algorithm}
Super Odometry is modeled using a factor graph. Here we mainly introduce three types of odometry factors for factor graph construction, namely: (a) IMU; (b) LiDAR-inertial; and (c) Visual-inertial odometry factors. 

\subsection{IMU Odometry Factors}
The structure of IMU odometry factors is shown in Fig.\ref{fig:algorithm}.({a}), which is similar to traditional pose graph structure. However, it is different since the estimated $state$ of IMU odometry not only contains position $\mathbf{p}_{b}^{w}$, velocity $\mathbf{v}_{b}^{w}$ and orientation 
$\mathbf{q}_{b}^{w}$ but also the accelerometer $\mathbf{b}_{a}$ and gyroscope bias $\mathbf{b}_{g}$. Each node in the pose graph is associated with a state. 
The density of nodes is determined by the lowest-frequency odometry. The edge between two consecutive nodes represents the relative body motion obtained from 
IMU preintegration method. Other edges could be local constraints or global constraints which depend on the type of sensors. 

\subsubsection{\textbf{IMU Preintegration Factor}}
Based on the IMU preintegration method proposed in \cite{forster2015imu}, we can obtain the relative motion measurement $\Delta \mathbf{r}=\left(\hat{\alpha}_{j}^{i}, \hat{\beta}_{j}^{i}, \hat{\gamma}_{j}^{{i}}\right)$ between consecutive IMU body frames $i$ and $j$. The IMU preintegration factor can be defined as:
\begin{equation}
	\label{eq:imu_factor}
	\begin{aligned}
		\mathbf{e}_{ij}^{{imu}}&=\mathbf{z}_{ij}^{b}-h_{ij}^{b}\left(\mathbf{x}_{i}, \mathbf{x}_{j}\right) \\
		&=\left[\begin{array}{c}
			\hat{\alpha}_{j}^{i}\\
			\hat{\beta}_{j}^{i} \\
			\hat{\gamma}_{j}^{i} \\
			\mathbf{b}_{a {j}} \\
			\mathbf{b}_{w {j}}\\ 
		\end{array}\right] \ominus\left[\begin{array}{c}
			\mathbf{R}_{w}^{i}\left(\mathbf{p}_{j}^{w}-\mathbf{p}_{i}^{w}+\frac{1}{2} \mathbf{g}\Delta t^{2}-\mathbf{v}_{i}^{w} \Delta t\right) \\
			\mathbf{R}_{w}^{i}\left(\mathbf{v}_{j}^{w}+\mathbf{g} \Delta t-\mathbf{v}_{i}^{w}\right)\\
			\mathbf{q}_{i}^{w_{}^{-1}} \otimes \mathbf{q}_{j}^{w_{}}\\
			\mathbf{b}_{a {i}}\\
			\mathbf{b}_{w {i}}\\
		\end{array}\right]
	\end{aligned}
\end{equation}

Where rotation matrices $\mathbf{R}$ and Hamilton quaternions $\mathbf{q} $ are used to represent rotation. $\mathbf{p}_{i}^{w}$, $\mathbf{v}_{i}^{w}$ and $\mathbf{q}_{i}^{w}$ are translation, velocity, and rotation from the body frame to the world frame. $\mathbf{g}$ is the gravity vector in the world frame.
$\mathbf{b}_{a}$ and $\mathbf{b}_{w}$ are accelerometer bias and gyroscope bias respectively. $\ominus$ is the minus operation on the IMU residuals.  
Note that the accelerometer bias error and gyroscope bias error are jointly optimized in the graph.   

\subsubsection{\textbf{Relative Pose Factor}}
Since the camera and LiDAR are not globally referenced and their odometry is based on the first pose of the robot, we only use their relative poses [$\Delta p_j^i$, $\Delta q_j^i$] as local constraints to constrain the IMU preintegration's position $\hat \alpha _j^i$, and rotation $\hat \gamma _j^i$. 
The relative pose factor $\mathbf{e}_{i j}^{LIO},\mathbf{e}_{i j}^{VIO}$ is derived as:

\begin{equation}
	\mathbf{e}_{i j}^{LIO},\mathbf{e}_{i j}^{VIO}=\left[ {\begin{array}{*{20}{c}}
{\Delta p_j^i}\\
{\Delta q_j^i}
\end{array}} \right] \ominus \left[ {\begin{array}{*{20}{c}}
{\hat \alpha _j^i}\\
{\hat \gamma _j^i}
\end{array}} \right]
\end{equation}

\subsubsection{\textbf{IMU Odometry Optimization}}
For each new frame, we minimize an optimization problem that consists of relative pose factor ($\mathbf{e}_{i j}^{LIO}$,$\mathbf{e}_{i j}^{VIO}$), and marginalization prior $\mathbf{E}_{m}$.
\begin{equation}
	E=\sum_{(i, j) \in \mathcal{B}} \mathbf{e}_{i j}^{LIO\top} {W}_{i j}^{-1} \mathbf{e}_{i j}^{LIO}+
	\sum_{(i, j) \in \mathcal{B}} \mathbf{e}_{i j}^{VIO\top}{W}_{i j}^{-1} \mathbf{e}_{i j}^{VIO}+\mathbf{E}_{\mathrm{m}}
\end{equation}
The relative pose factor has to be weighted with the appropriate covariance matrix ${W}_{ij}$, which can be calculated based on the reliability of the observation. For example, in visually degraded environments, $\mathbf{e}_{ij}^{VIO}$ calculated by VIO will have a lower weight. In contrast, in geometrically degraded environments, $\mathbf{e}_{ij}^{LIO}$ calculated by LIO will have a lower weight. Note that without loss of generality, the IMU odometry can also incorporate measurements from other sensors such as GPS and wheel odometry. 


\subsection{LiDAR-Inertial Odometry Factors}
The structure of LiDAR-Inertial odometry factors is shown in Fig.\ref{fig:algorithm}.({c}). Since IMU odometry is computationally efficient and outputs at a very high frequency, its IMU preintegration factor can be naturally added into the factor graph of LiDAR-Inertial odometry. The IMU preintegration factor will be used as motion prediction for current scan-map matching and connects consecutive LiDAR frames in the factor graph. 


The pipeline of the scan-map matching process can be divided into four steps, namely: (a) PCA-based Feature Extraction (b) Multi-metric Linear Square ICP (c) LiDAR-inertial Odometry Factor (d) Dynamic Octree.

\subsubsection{\textbf{PCA Based Feature Extraction}}
When the system receives a new LiDAR scan, the LiDAR scan will be downsampled to a fixed number and then fed into the feature extraction module. The point-wise K-D tree is used to find the nearest $k$ neighbor points within a sphere of radius $r$. Based on these neighbor points, the principal components analysis (PCA) method is used to analyze the local linearity $\sigma_{1 \mathrm{D}}$, planarity $\sigma_{2 \mathrm{D}}$, and curvatures $\sigma_{3 \mathrm{D}}$ of geometrical feature. They are defined as\cite{hackel2016fast}
\begin{equation}
	\sigma_{1 \mathrm{D}}=\frac{\sigma_{1}-\sigma_{2}}{\sigma_{1}}, \sigma_{2 \mathrm{D}}=\frac{\sigma_{2}-\sigma_{3}}{\sigma_{1}}, \sigma_{3\mathrm{D}}=\frac{\sigma_{3}}{\sigma_{1}+\sigma_{2}+\sigma_{3}}
	\label{eq:feature}
\end{equation}
where $\sigma_{i}=\sqrt{\lambda_{i}}$ and $\lambda_{i}$ are the eigenvalues of the PCA for the normal computation. According to the magnitude of local feature $\sigma_{1 \mathrm{D}}$,$\sigma_{2 \mathrm{D}}$, $\sigma_{3 \mathrm{D}}$, point, line and plane features $\mathbb{F}_{i+1}=\{F^{po}_{i+1}, F^{li}_{i+1},F^{pl}_{i+1}\}$ can be easily classified.

\subsubsection{\textbf{Multi-metric ICP Factors}}
After extracting reliable geometric features,
we use motion prediction provided by IMU odometry and transform the $\mathbb{F}_{i+1}$ from $\bold{B}$ to $\bold{W}$. Then, we find its point, line, and plane correspondence in the map by using the proposed dynamic octree. For the sake of brevity, the detailed procedures will be described in the next section.

To improve the robustness of Super Odometry, we expect to estimate the optimal transformation that jointly minimizes the point-to-point, point-to-line, and point-to-plane distance error metric, which can be computed using the following equations:
\begin{equation}
	\begin{cases}
		\mathbf{e}_{i}^{\mathrm{po} \rightarrow \mathrm{po}}=\mathbf{q}_{i}-(R\mathbf{p}_{i}+t) & \text{$q_{i},p_{i}\in F^{po}_{i+1}$}
		\\
		\mathbf{e}_{j}^{\mathrm{po} \rightarrow \mathrm{li}}=\mathbf{v}_{j} \times\left(\mathbf{q}_{j}-(R\mathbf{p}_{j}+t)\right)
		& \text{$q_{j},p_{j}\in F^{li}_{i+1}$}
		\\
		\mathbf{e}_{k}^{\mathrm{po} \rightarrow \mathrm{pl}}=\mathbf{n}_{k}^{\top}\left(\mathbf{q}_{k}- (R\mathbf{p}_{k}+t)\right) 
		& \text{$q_{k},p_{k}\in F^{pl}_{i+1}$}
	\end{cases}
\end{equation}

where $\mathbf{e}_{i}^{\mathrm{po} \rightarrow \mathrm{po}}$, $	\mathbf{e}_{j}^{\mathrm{po} \rightarrow \mathrm{li}}$ and $\mathbf{e}_{k}^{\mathrm{po} \rightarrow \mathrm{pl}}$ are the point-to-point (line, plane) distance. $\mathbf{v}_{j}$ and $\mathbf{n}_{k}$ represent the corresponding principal and normal direction of features. $\mathbf{T}_{i+1}=\{\bold{R}, \bold{t}\}$ is the optimal transformation we expect to estimate. 

However, in airborne obscurants environments such as dust, fog, and smoke environments, we may extract geometric features on these noise data and obtain unreliable data association. To solve this problem, we found that the quality of correspondence is depended on if the extracted features' neighbor points fit the point, line, or plane distribution. Therefore, we use distribution's quality ${W}_{l}=\{w_i^{po \to po}, w_i^{po \to li},w_i^{po \to pl}\}$ to define the quality of correspondence by using the following equation.  

\begin{equation}
\label{eq:w1}
{w_i^{po \to po} = \frac{{{\sigma _{3D}}}}{{{\sigma _{3Dmax}}}}}
\end{equation}

\begin{equation}
\label{eq:w2}
{w_i^{po \to li/pl} = 1 - \frac{{\sum\limits_{i = 1}^k {{{\left( {{p_i} - \bar p} \right)}^T}A\left( {{p_i} - \bar p} \right)} }}{{k \times {d_{max}}}}}
\end{equation}

where $\sigma_{3Dmax}$ is the maximum threshold of curvatures.
$p_{i}$ are the neighbor points of extracted features and $\bar p$ is their mean. For point-to-line(plane) distance error, we use normal direction $\mathbf{n}_{k}$ of features and define $A=I-\mathbf{n}_{k}^T \times \mathbf{n}_{k}$ and $A=\mathbf{n}_{k}^T \times \mathbf{n}_{k}$ respectively.
$d_{max}$ represents the maximum distance error for point-to-line(plane) correspondence. $k$ is the number of neighbor points.  

\subsubsection{\textbf{LiDAR-inertial Odometry Optimization}} The levenberg-marquardt method is then used to solve for the optimal transformation by minimizing:

\begin{equation}
\mathop {\min }\limits_{{\mathbf{T}_{i + 1}}} \left\{ {\left. {\begin{array}{*{20}{c}}
{\sum\limits_{p \in {\mathbb{F}_{i}}} {{{\left\| {{W_l}e_i^{po \to po,li,pl}} \right\|}^2} + } }\\
{\sum\limits_{\left( {i,i + 1} \right) \in \mathcal{B} } {{{\left\| {{W_{imu}}{e^{imu}}} \right\|}^2} + \mathbf{E}_{imuodom}^{prior}} }
\end{array}} \right\}} \right.
\end{equation}

$\mathbf{E}_{imuodom}^{prior}$  and $\mathbf{e}_{}^{imu}$ are the predicted pose prior and IMU preintegration factor provided by IMU odometry. $\mathbf{W}_{l}$ and $\mathbf{W}_{imu}$ are the weight of LiDAR and imu correspondence.  

When the environment is geometrically degraded, since the IMU odometry can receive the measurements from other sensors, the IMU odometry is still reliable and $E_{imuodom}^{prior}$ will be dominant in the optimization problem. Meanwhile, unreliable LiDAR constraints will be rejected or have low weight by evaluating the quality of constraints. On the other hand, when the environment is well-structured, the multi-metric ICP factor will be dominant in the optimization problem.

\subsubsection{\textbf{Dynamic Octree}}
\begin{figure}
	\centering
	\includegraphics[width=0.85\columnwidth]{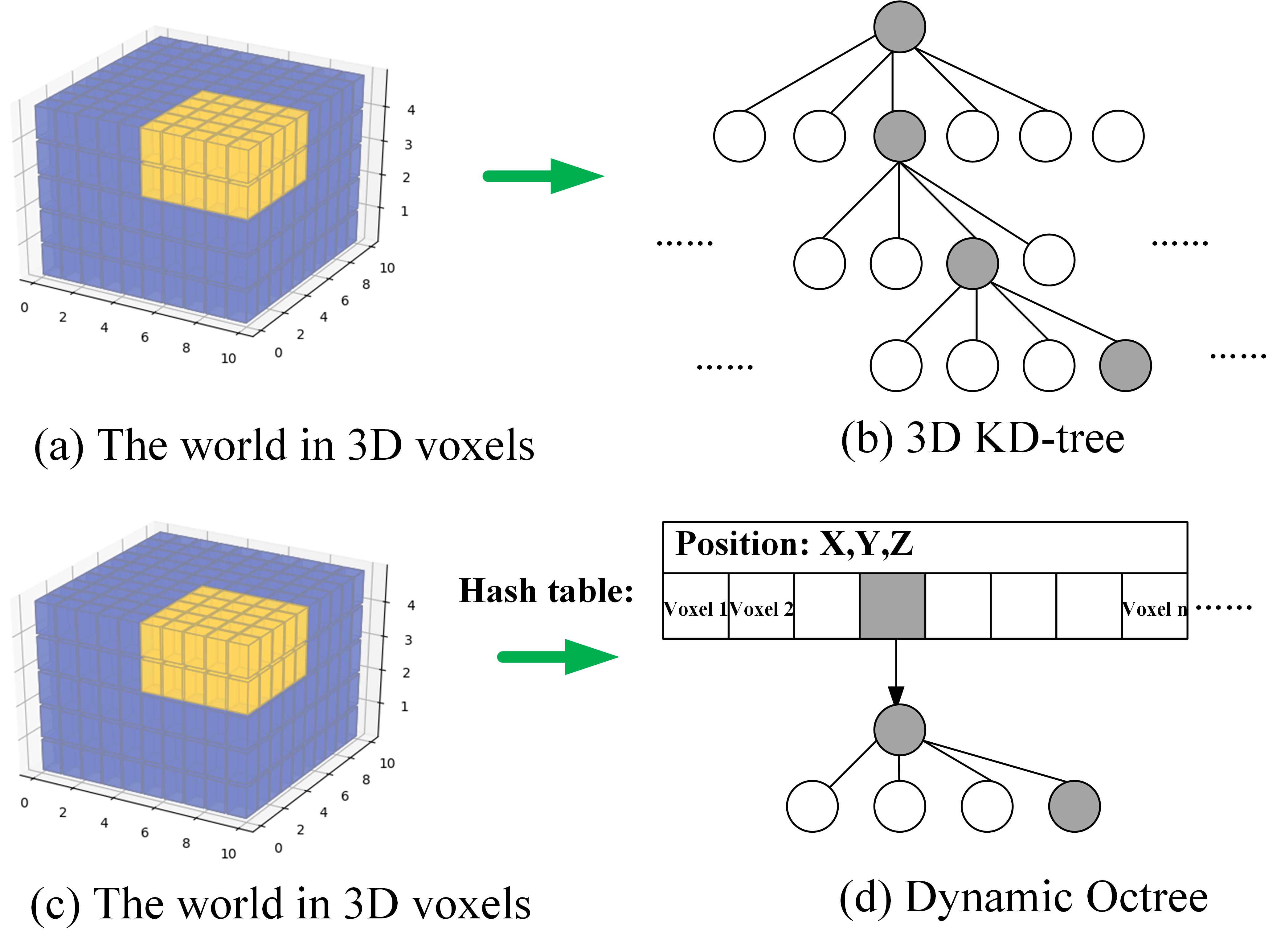}
	\caption{The comparison of 3D KD-tree and dynamic octree. (a,b) shows the construction process of 3D KD-tree when adding new point cloud (grey circle), which needs to change the structure of the whole tree. (c,d) shows the construction process of dynamic octree when adding new point cloud (grey circle), which only needs to change the structure of the sub-tree.}
	\vspace{-3mm}
	\label{fig:dynamic_octree}
\end{figure}

Most LiDAR SLAM methods adopt the KD-tree method to organize the 3D points and fulfill data association. However, we argue that the traditional KD-tree is very time-consuming to organize the 3D points since it only uses one tree to organize all the points and needs to recreate the KD-tree every time when adding new points shown in Fig.\ref{fig:dynamic_octree}(b). 
  To solve this problem, we propose to use  a more efficient organization of 3D points named dynamic octree to facilitate data association. The dynamic octree is based on the work of Behley \cite{behley2015icra}. It stores the map as a hash table as shown in Fig.\ref{fig:dynamic_octree}(d). 
The world is represented by voxels and voxels can be accessed from the hash table by using their XYZ indices. Instead of only building one tree, each voxel will have its octree to organize the points and each octree can be accessed by a hash table. Since we only need to update the specific octree instead of the whole tree, the real-time performance on data association will be significantly improved.


	
			


\subsection{Visual-Inertial Odometry Factors}
To take full advantage of the fusion of vision and LiDAR sensing modalities, we track monocular visual features and LiDAR points within the camera field of view and use them to provide depth information for visual features.

\subsubsection{ \textbf{Visual-inertial Odometry Optimization}}
For each new keyframe, we minimize a non-linear optimization problem that consists of visual reprojection factor $\mathbf{e}_{reproj}$, IMU preintegration factor $\mathbf{e}_{imu}$, marginalization factor $\mathbf{E}_{m}$ and a pose prior from IMU odometry $\mathbf{E}_{imuodom}^{prior}$. The structure of Visual-inertial odometry factors is shown in Fig.\ref{fig:algorithm}(b).

\begin{equation}
	\min _{\mathbf{T}_{i+1}}\left\{
	\begin{split}
		&\sum_{ i \in \mathbf{obs}(i)}\mathbf{e}_{\text {reproj }}^{\mathbf{T}} W_{\text {reproj}}^{-1} \mathbf{e}_{\text {reproj }}^{\mathbf{}}+\sum_{(a, b) \in \mathbf{C}}\mathbf{e}_{\text {imu}}^{\mathbf{T}} W_{imu}^{-1} \mathbf{e}_{imu}^{\mathbf{}}
		\\ 
		&+\mathbf{E}_{m} + \mathbf{E}_{imuodom}^{prior}	
	\end{split}
	\right\}
\end{equation}

$\mathbf{obs(i)}$ represents a set that contains visual features $i$ that is tracked by other frames. The set $\mathbf{C}$ contains pairs of visual nodes $(a, b)$ connected by IMU factors. $W_{\text {reproj}}$ and $W_{\text {imu}} $ are covariance matrices for visual reprojection factor and IMU preintegration factor. 

Similar to the explanation in the last section, when the environment is visually degraded, the $\mathbf{E}_{imuodom}^{prior}$ will be dominant in the optimization problem and unreliable vision factor will be rejected by analyzing the information matrix. However, when the environment has good lighting conditions, the visual factor will be dominant in the optimization problem and the IMU preintegration factor only provides the initial guess for visual feature tracking.

\section{Experiments}
 In this section, we evaluate the robustness of Super Odometry with other state-of-the-art algorithms to degenerate sensor inputs. Then, we evaluate the real-time performance of the algorithms. Our experiment video can be found through this link: \href{https://sites.google.com/view/superodometry}{https://sites.google.com/view/superodometry}. 

\subsection{Dataset}
\begin{figure}[h]
	\centering
 	\includegraphics[width=1.0\columnwidth]{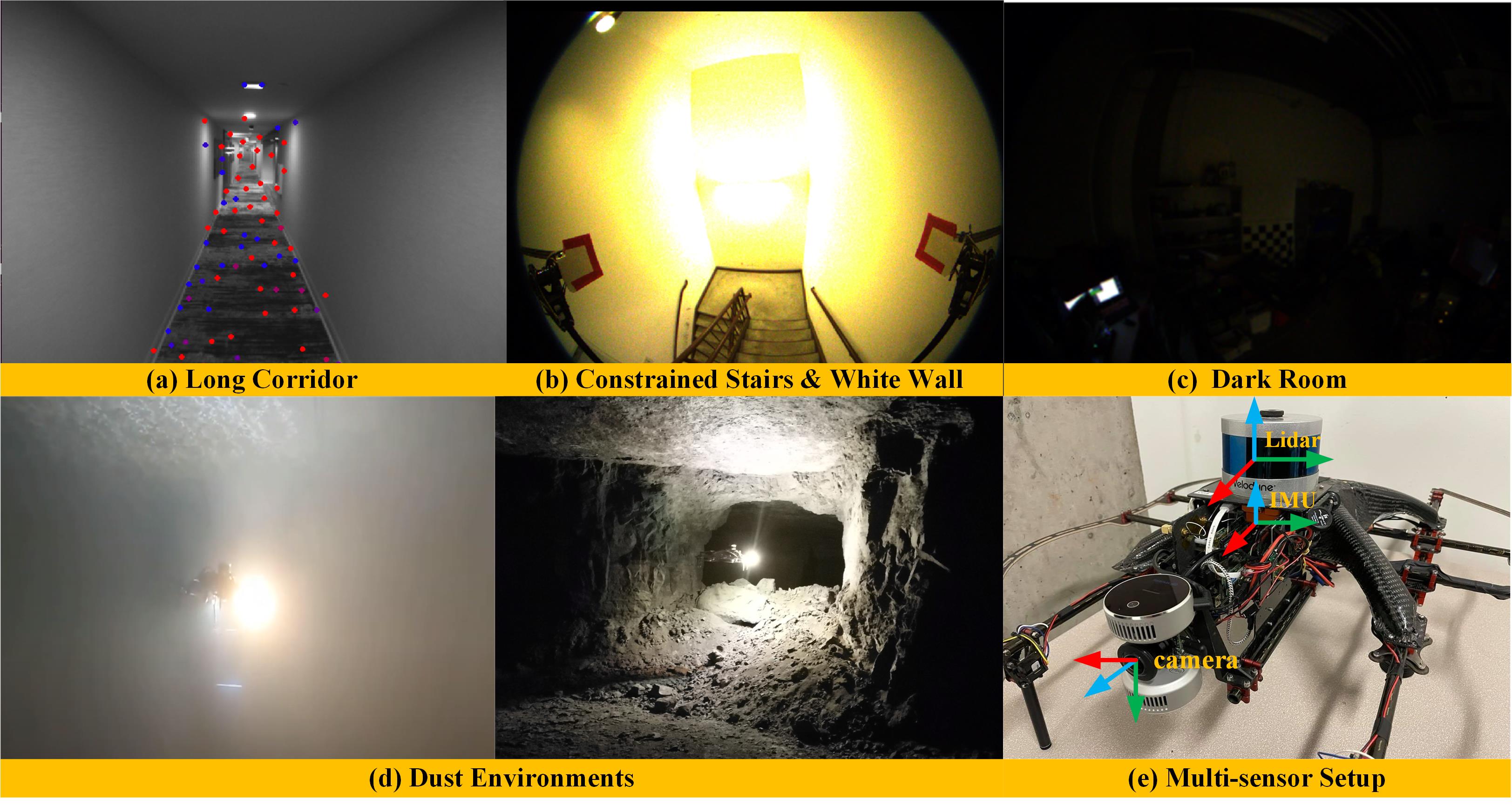}
	\caption{(a-d) shows the challenging environments where we collect datasets. (e) shows our multi-sensor setup. }
	\vspace{-3mm}
	\label{fig:real_scene}
\end{figure}

We collected our test dataset with Team Explorer's DS drones (Fig.\ref{fig:real_scene}(e)), deployed in the DARPA Subterranean Challenge. It has a multi-sensor setup
including a Velodyne VLP-16 LiDAR, an Xsens IMU, a uEye camera with a wide-angle fisheye lens, and an Intel NUC onboard computer. The data sequences were designed to include both visually and geometrically degraded scenarios, which are particularly troublesome for camera- and LiDAR-based state-estimation.

\begin{figure*}[ht]
	\centering
	\subfigure[VINS]{\includegraphics[width=.195\textwidth]{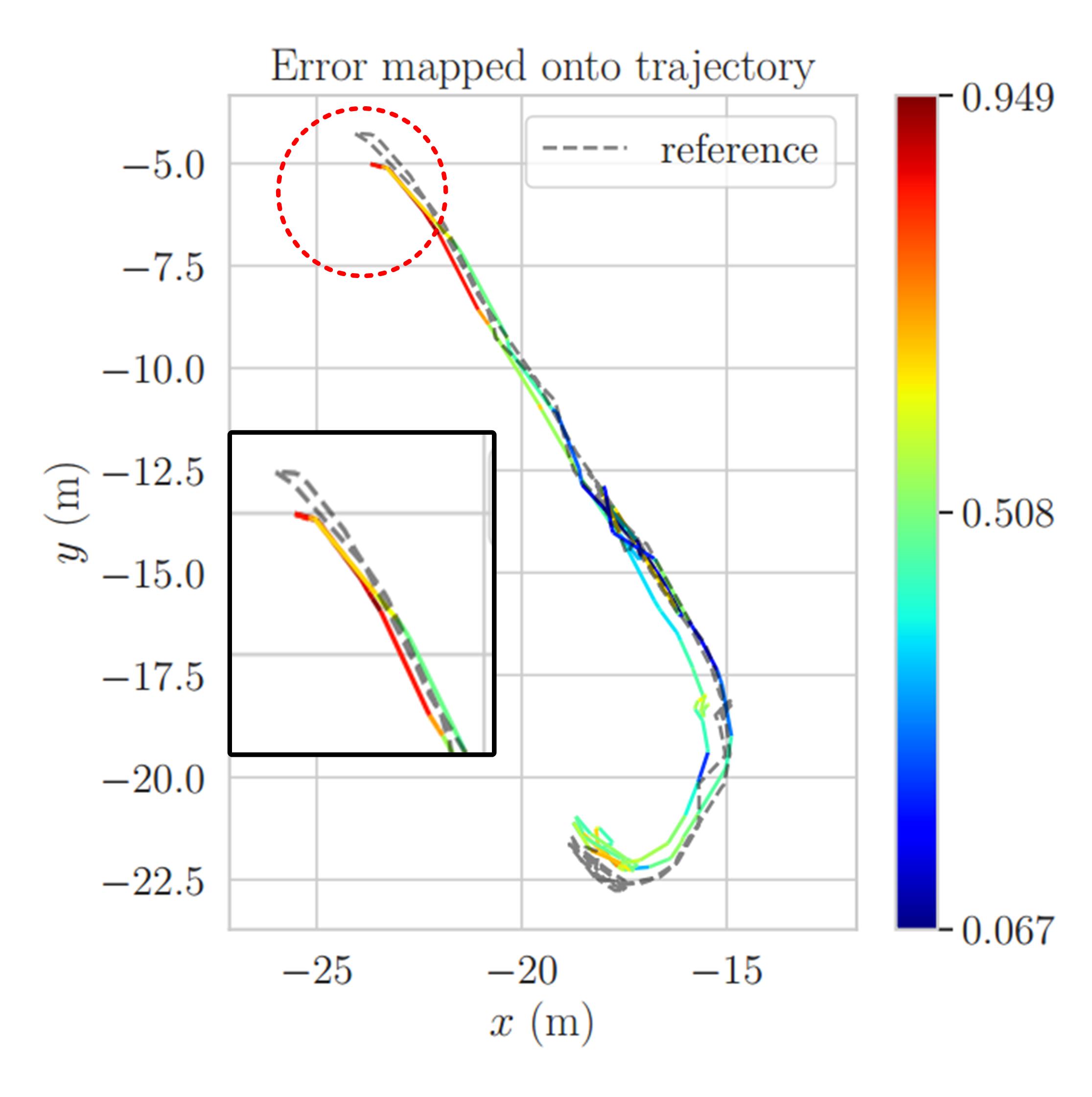}}
	\subfigure[VINS-Depth]{\includegraphics[width=.195\textwidth]{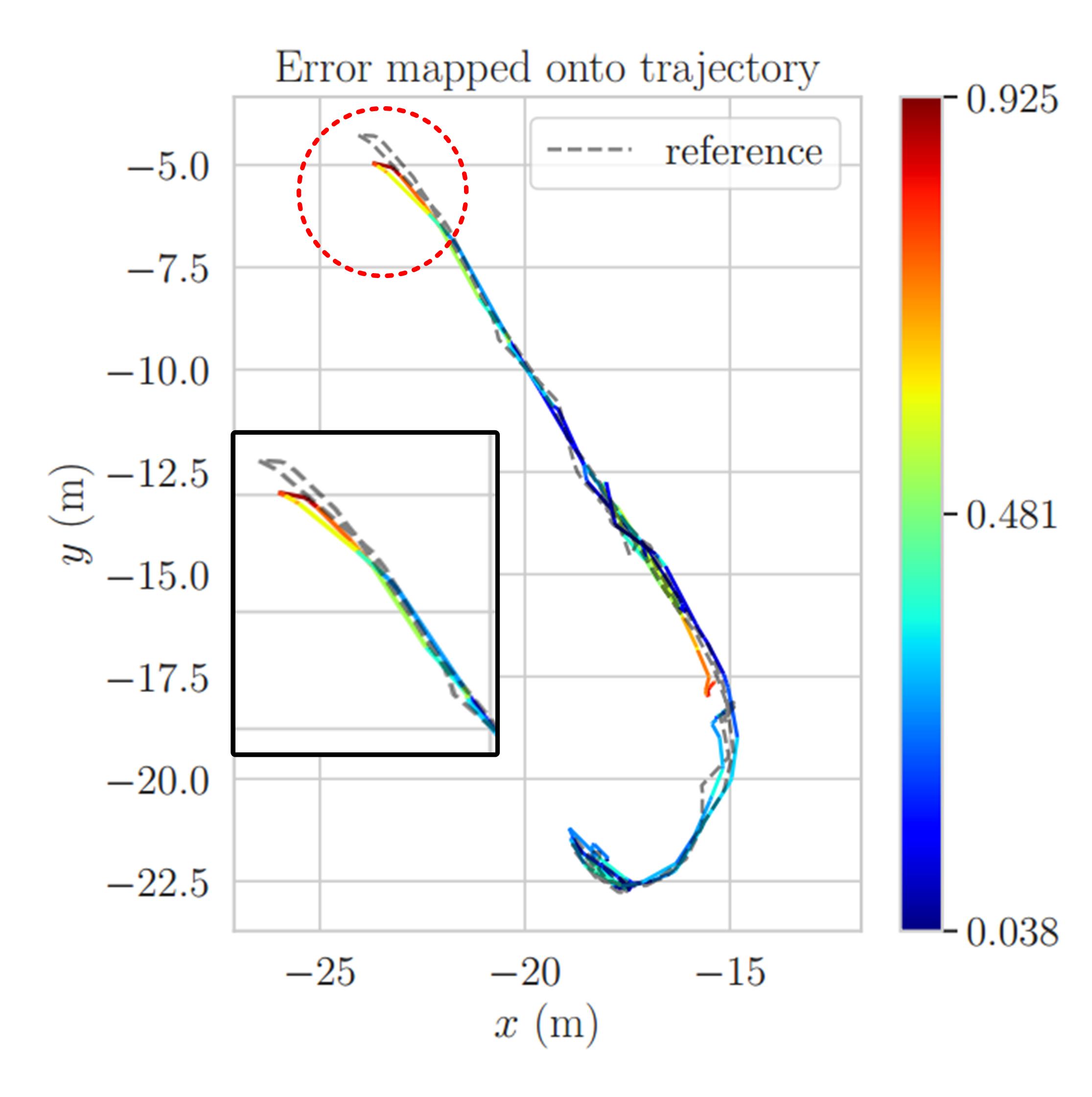}}
	\subfigure[LOAM]{\includegraphics[width=.195\textwidth]{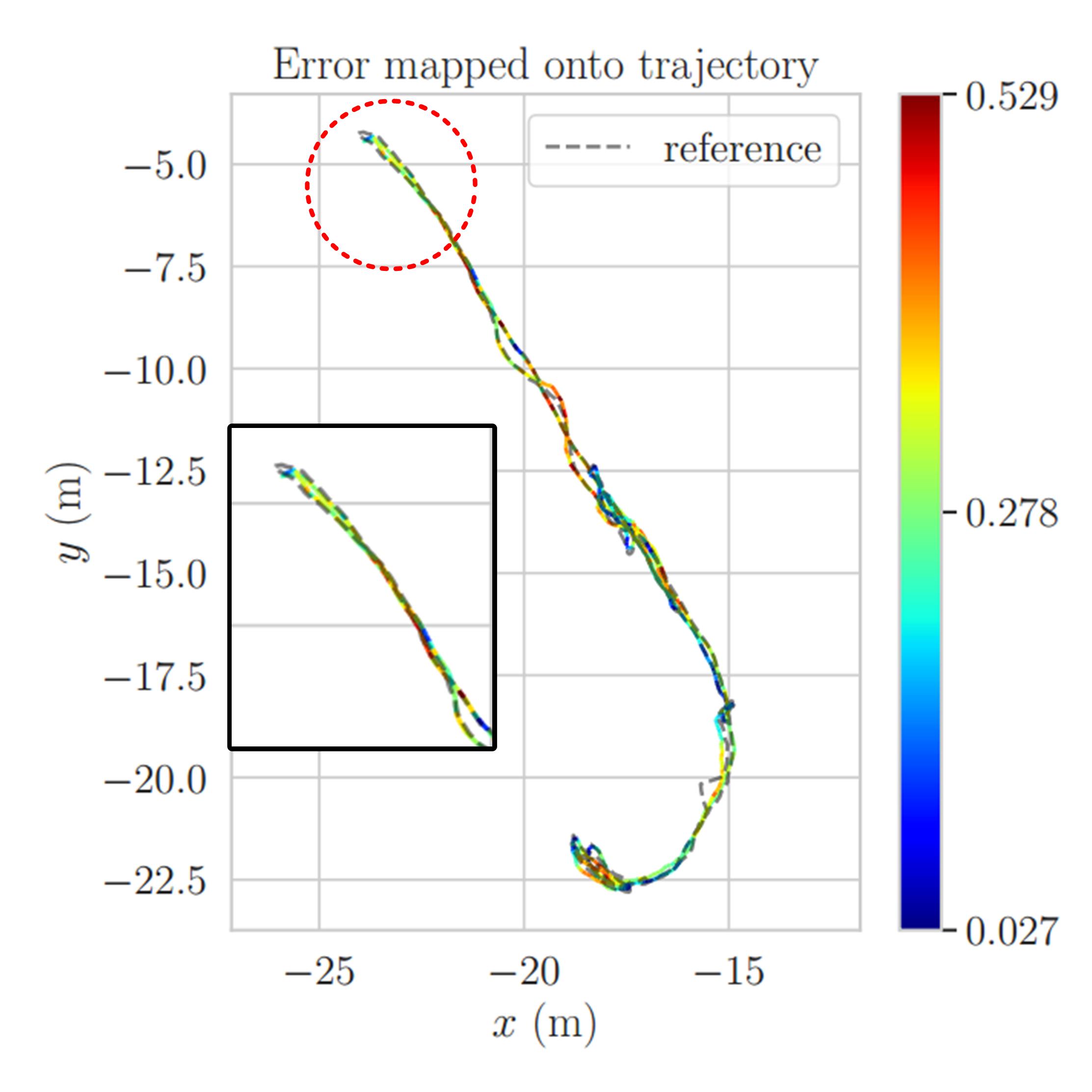}}
	\subfigure[LIO-SAM]{\includegraphics[width=.195\textwidth]{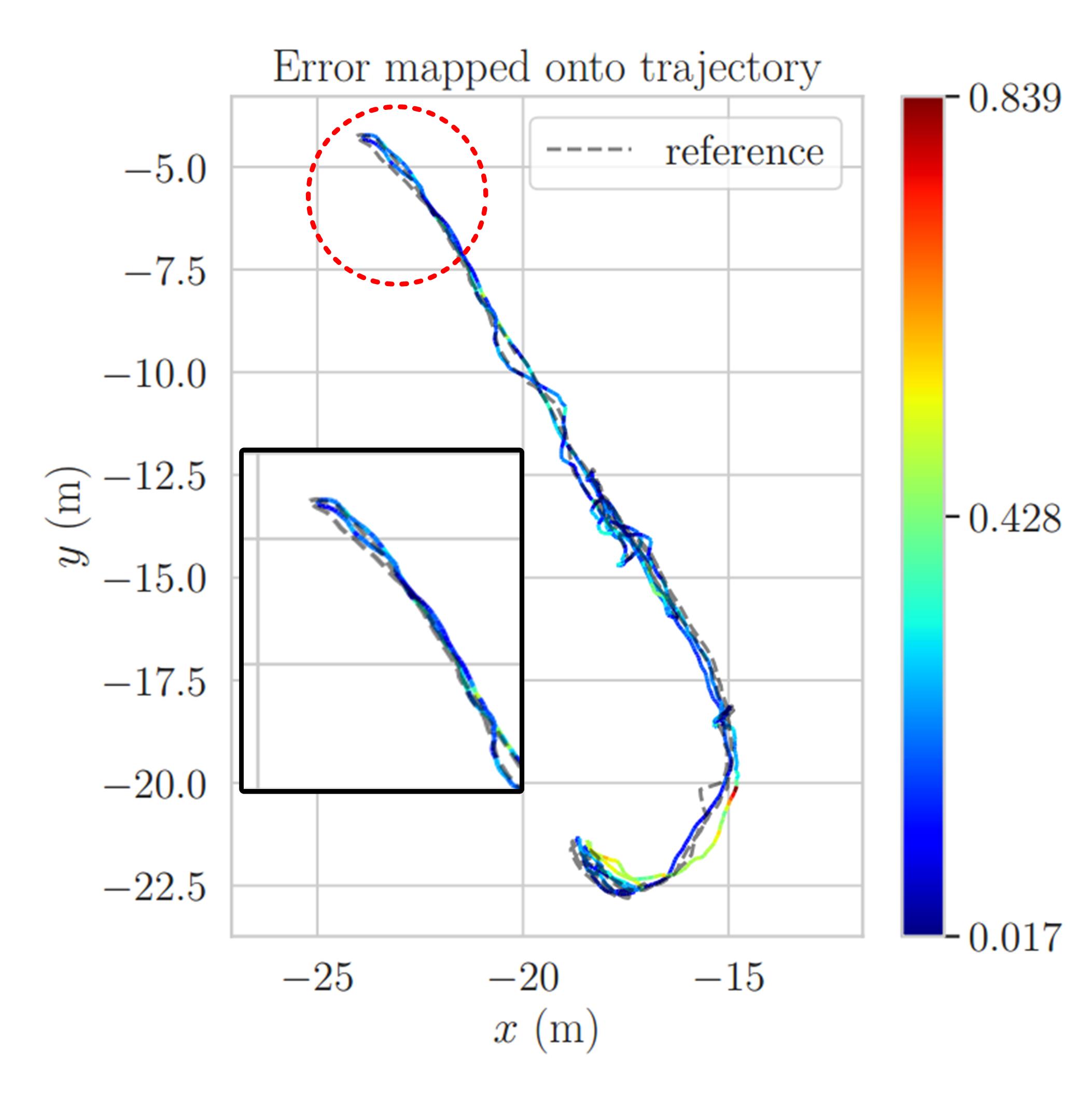}}
	\subfigure[Super Odometry]{\includegraphics[width=.195\textwidth]{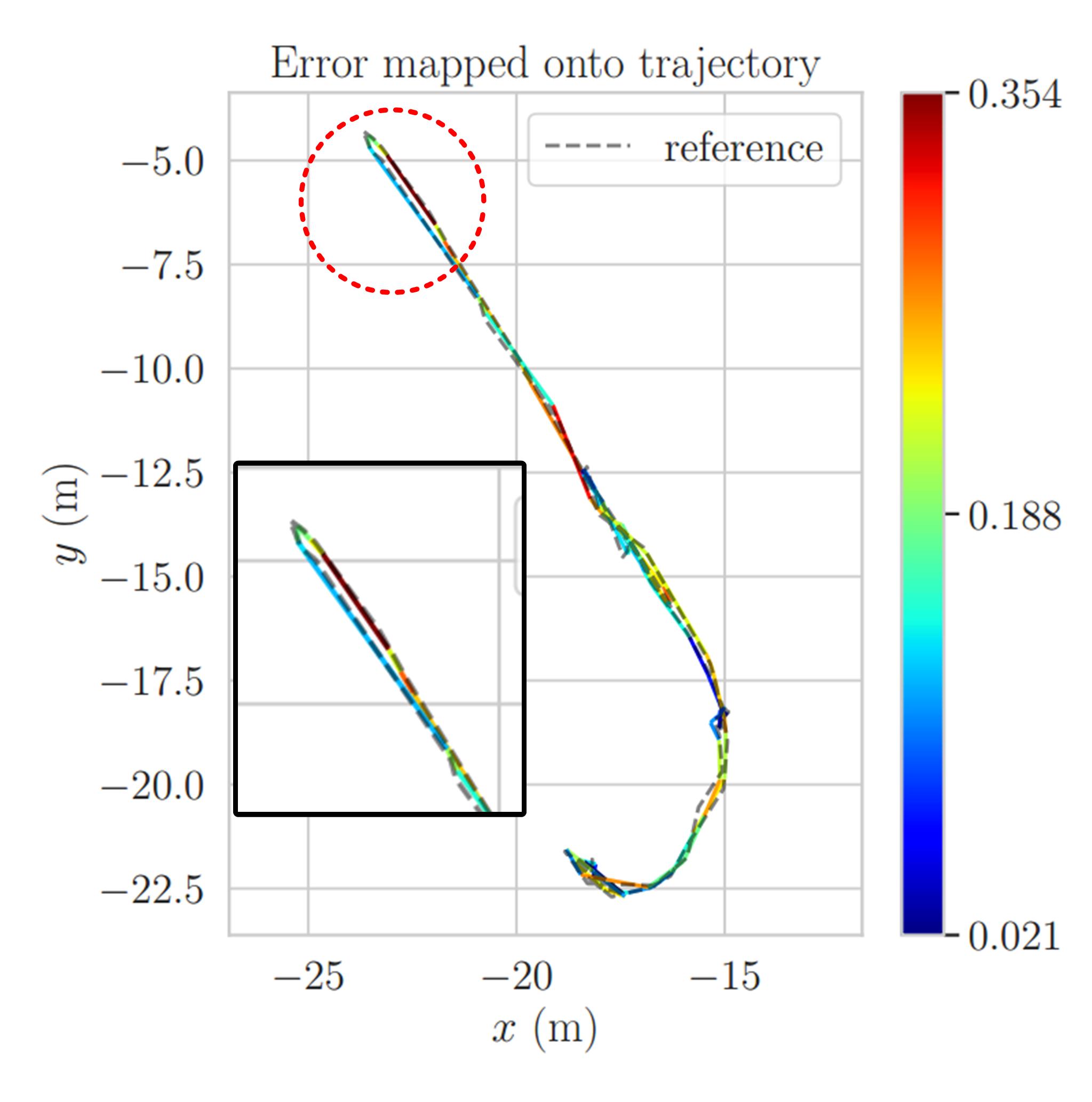}}
	\caption{Trajectories of LOAM, LIO-SAM, VINS, Depth-enhanced VINS, and Super Odometry on dark room dataset}
	\label{fig:lowlight_traj}
	\vspace{-2mm}
\end{figure*}

 The list of test sequences is as follows:
\begin{itemize}
	\item \textbf{Dark-Room}: Hand-carried walk in and out of a dark room shown as Fig.\ref{fig:real_scene}(c).
	\item \textbf{White-Wall}: Hand-carried walk with camera facing to a feature-less white wall shown as Fig.\ref{fig:real_scene}(b).
	\item \textbf{Constrained-Environment}: Hand-carried walk in a stair shaft shown as Fig.\ref{fig:real_scene}(b). 
	
	
	\item \textbf{Long-Corridor}: Hand-carried walk through a long corridor in an apartment building shown as Fig.\ref{fig:real_scene}(a). 
	\item \textbf{Dust}: Autonomous flying through a subterranean cave with heavy dust shown as Fig.\ref{fig:real_scene}(d).
	\item \textbf{Urban Challenge Alpha Course}: Competition run in DARPA SubT Urban Chellange Alpha Course in an abandoned nuclear facility shown as Fig.\ref{fig:alpha_course}.
\end{itemize}


\begin{figure*}[ht]
	\centering
	\subfigure[LOAM]{\includegraphics[width=.325\textwidth]{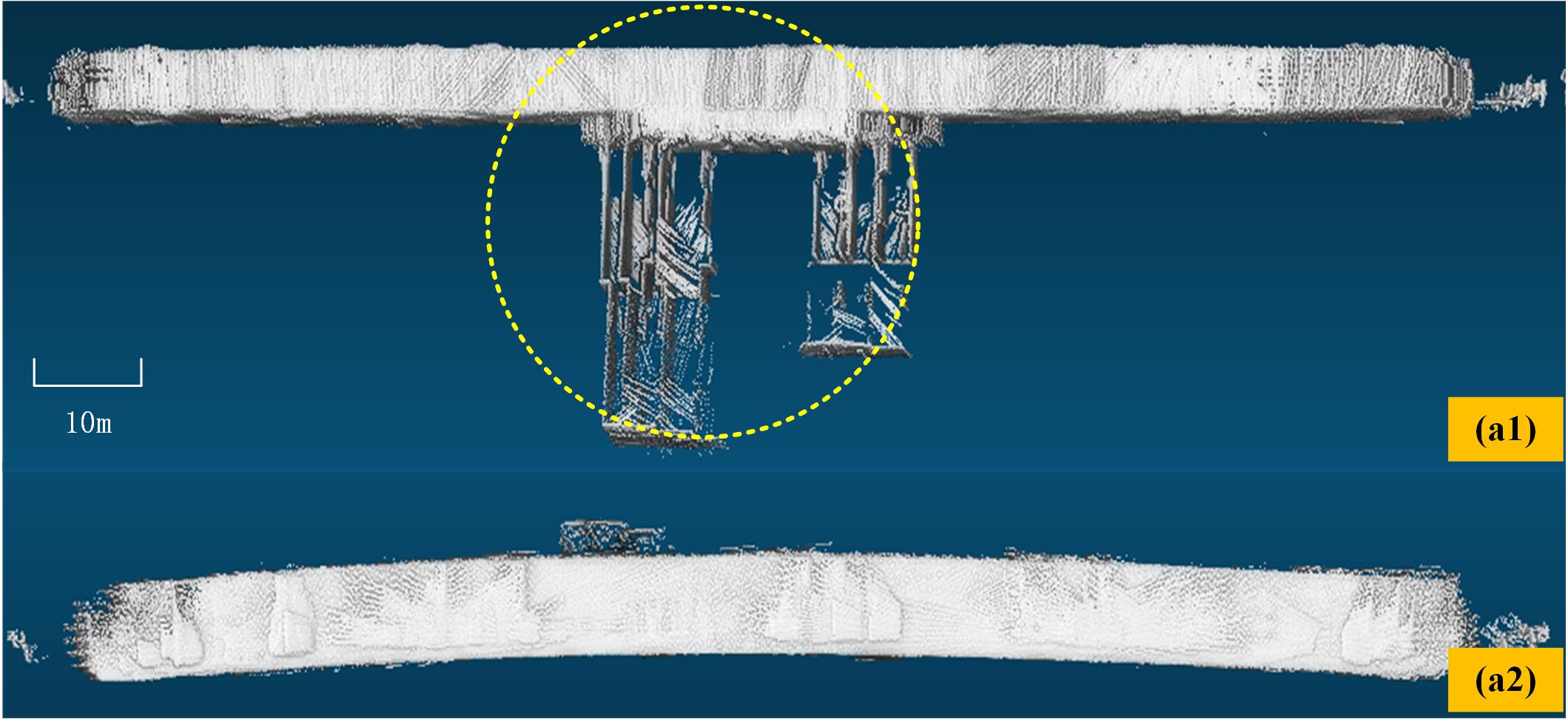}}
	\subfigure[LIO-SAM]{\includegraphics[width=.330\textwidth]{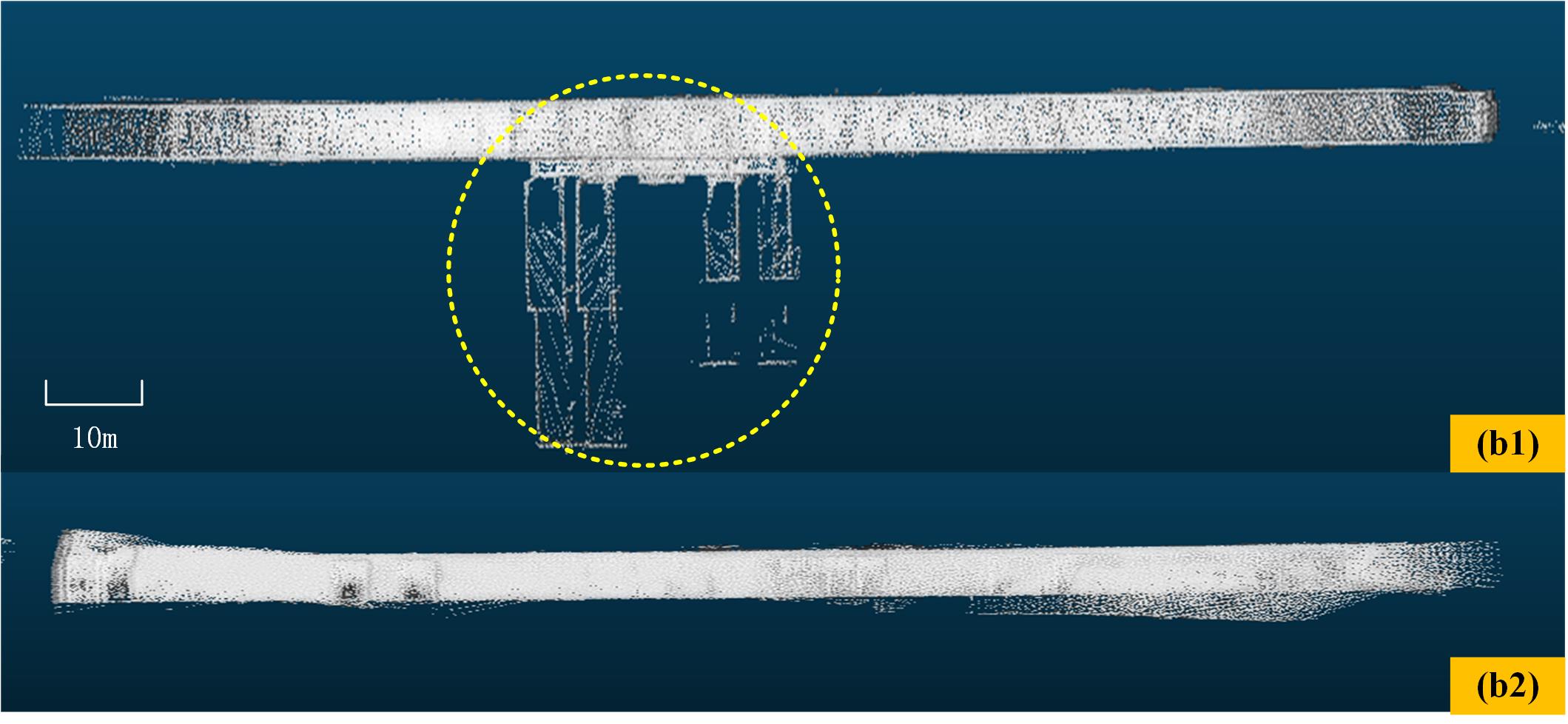}}
	\subfigure[Super Odometry]{\includegraphics[width=.325\textwidth]{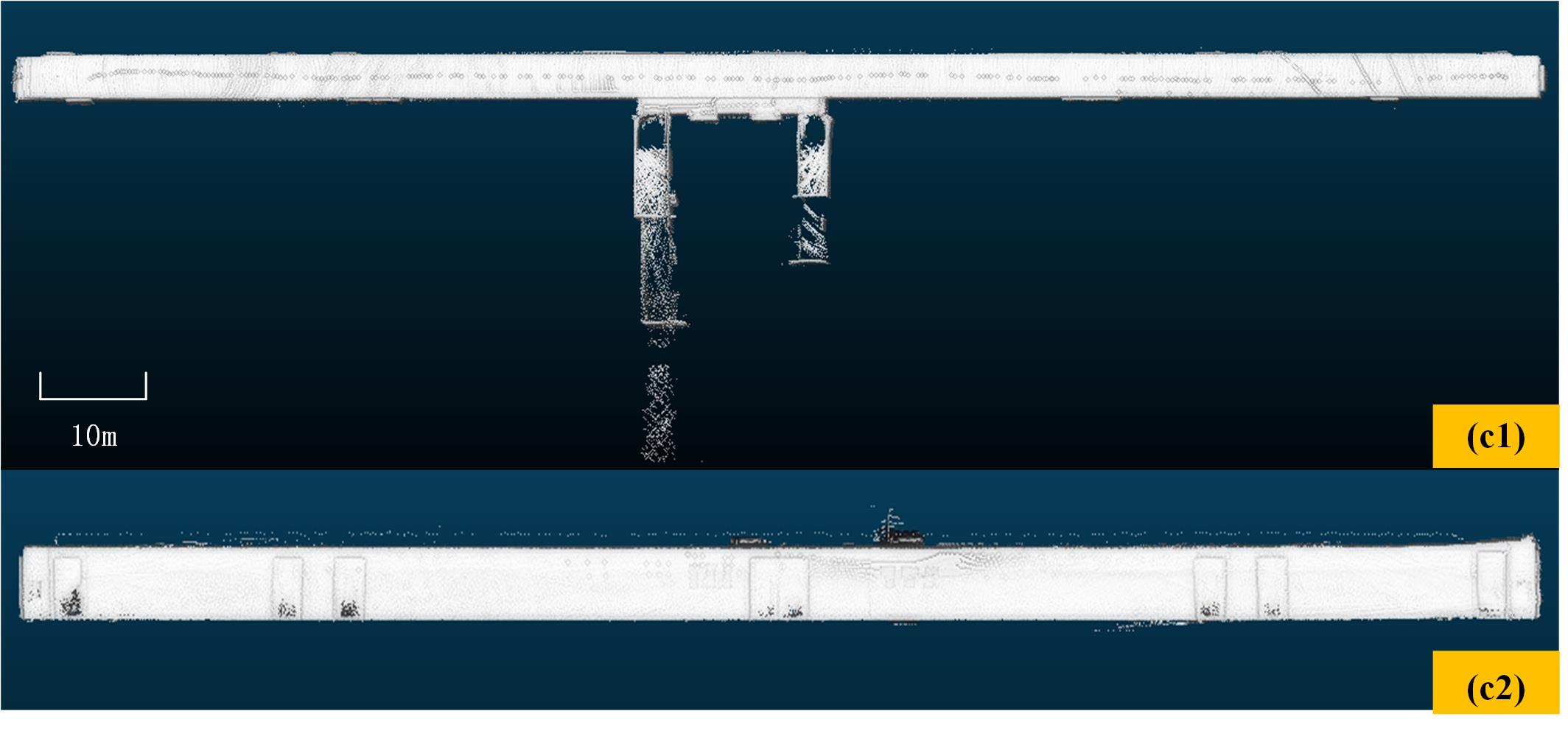}}
	\caption{Map comparison of LIO-SAM, LOAM, and Super Odometry in a long corridor data sequence. Upper image shows a top-down view and lower 
		image shows the z-drift of the result map. We can see that Super Odometry is free of misalignment and gives a higher quality map.}
	\label{fig:long-corridor-map}
	\vspace{-2mm}
\end{figure*}

The trajectory ground truth for quantitative analysis on \textbf{Dark-Room}, \textbf{Long-Corridor}, \textbf{White-Wall}, and 
\textbf{Constrained-Environment} is generated with a Total Station (TS) and tracking prism on top of the sensor suite.
Data sequences \textbf{Dust} and \textbf{Urban Challenge Alpha Course} do not have trajectory ground truth and are used for map quality and 
real-time performance analysis.

\subsection{Robustness and Accuracy Evaluation}
Our dataset contains visually and geometrically degraded scenes. Such conditions are common in underground environments, such as the ones encountered in each phase of the DARPA Subterranean Challenge. We compare our method with other LiDAR-based odometry (LOAM \cite{loam}, LIO-SAM\cite{shan2020liosam}) and vision-based odometry (VINS \cite{VINS}, Our Depth-enhanced VINS\cite{vinsicra}). We adopt EVO \footnote{https://github.com/MichaelGrupp/evo}  package to calculate the translational Absolute Trajectory Error (ATE) of each estimated trajectory against the ground truth trajectories. The maximum and RMSE results of ATEs are shown in Table \ref{tab:accuracy}. It is worth mentioning that no loop closing step is used in the experiments. We focus on comparing only the odometry part of the 
algorithms.

First, we consider visually-degraded environments, where camera-based algorithms are expected to give poor results. Next, we consider geometrically-degraded environments, where LiDAR-based algorithms usually suffer. Finally, we consider both types of degradation together. We show that Super Odometry has better accuracy in all cases, because of its robustness properties.

\subsubsection{\textbf{Robustness Comparison in Visually Degraded Environments}} 
Here we show the results of the \textbf{Dark-Room} sequence. In this data sequence,
we hand-carried the drone and walk past a poorly lit room combined with aggressive motion. The image is mostly dark (Fig.\ref{fig:real_scene}.(c)) and therefore challenging for feature extraction in the front-end of visual odometry.

Fig.\ref{fig:lowlight_traj} shows the algorithm output trajectories compared with the ground truth trajectory from the Total Station. Considering the 
the absolute trajectory error (ATE) shown in Table \ref{tab:accuracy}, we see that purely vision-based methods have the worst performance in this situation, as expected. 
Our IMU-centric Super Odometry outperforms the above methods with the lowest ATE of $0.174$m.
We also evaluate our method on the \textbf{White-Wall} data sequence mentioned in the dataset section. The ATE of each method can be found in Table \ref{tab:accuracy}. For more details, we refer readers to the accompanying video.


\begin{figure*}[ht]
	\centering
	\subfigure[LOAM]{\includegraphics[width=.34\textwidth]{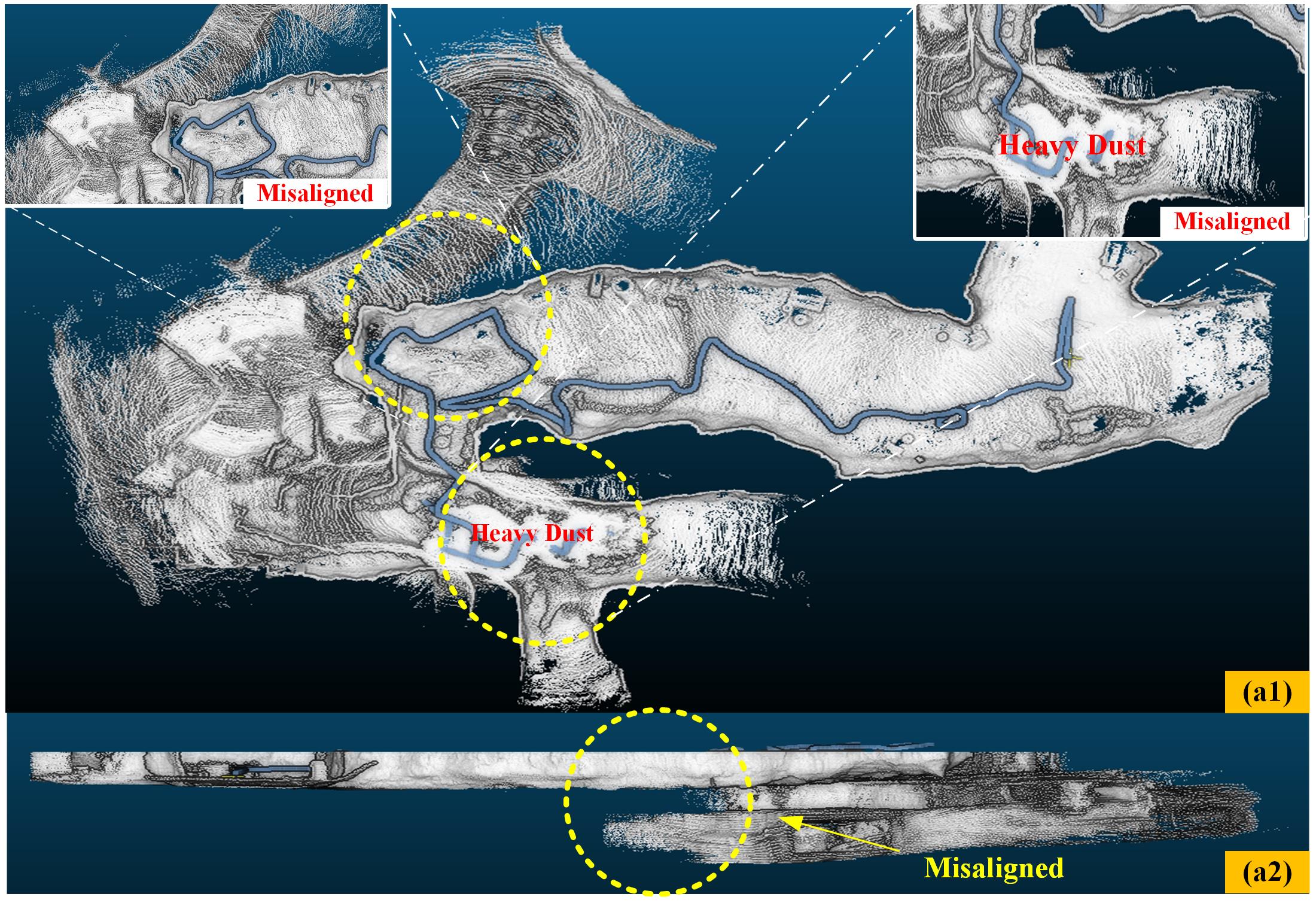}}
	\subfigure[LIO-SAM]{\includegraphics[width=.34\textwidth]{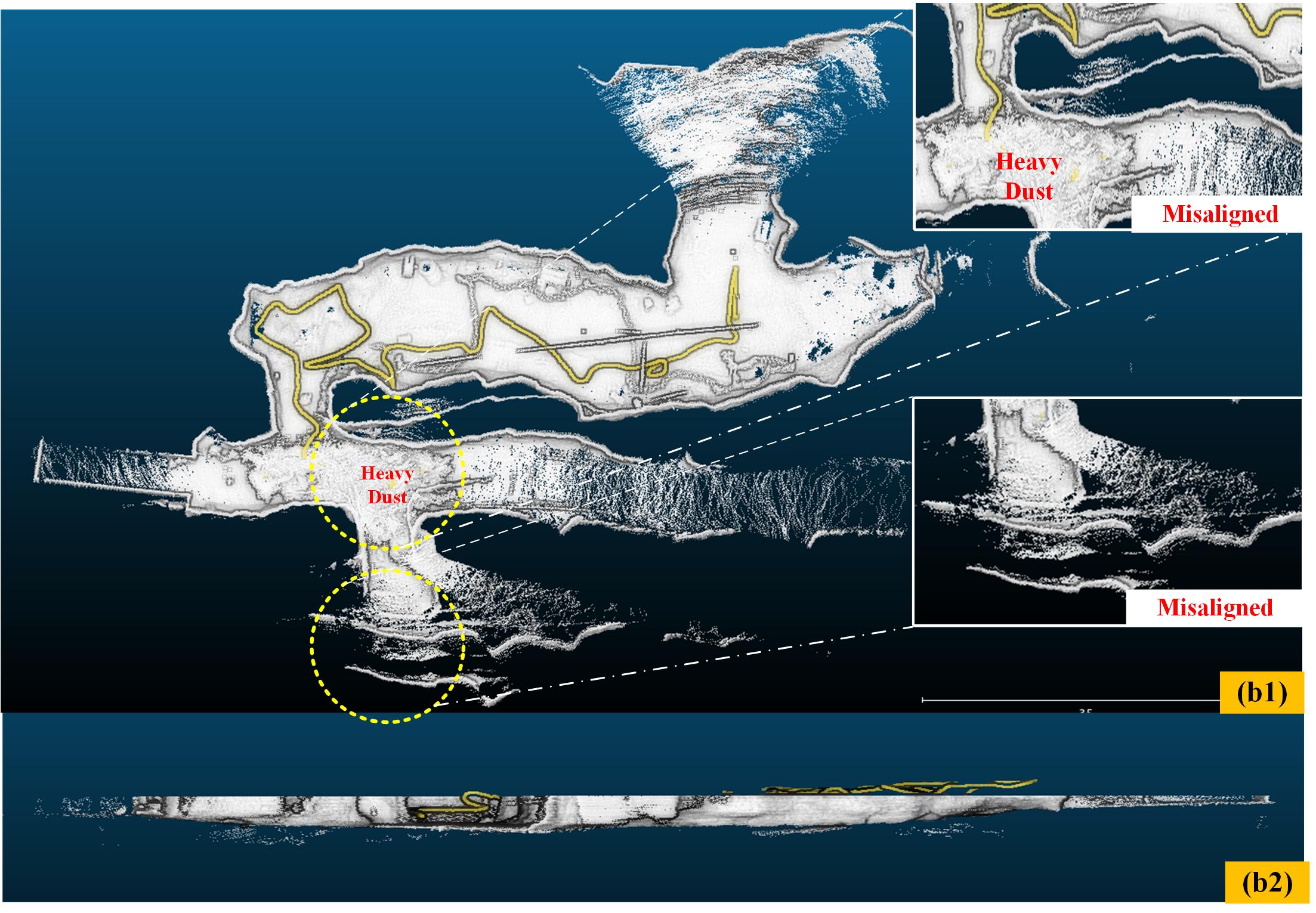}}
	\subfigure[Super Odometry]{\includegraphics[width=.275\textwidth]{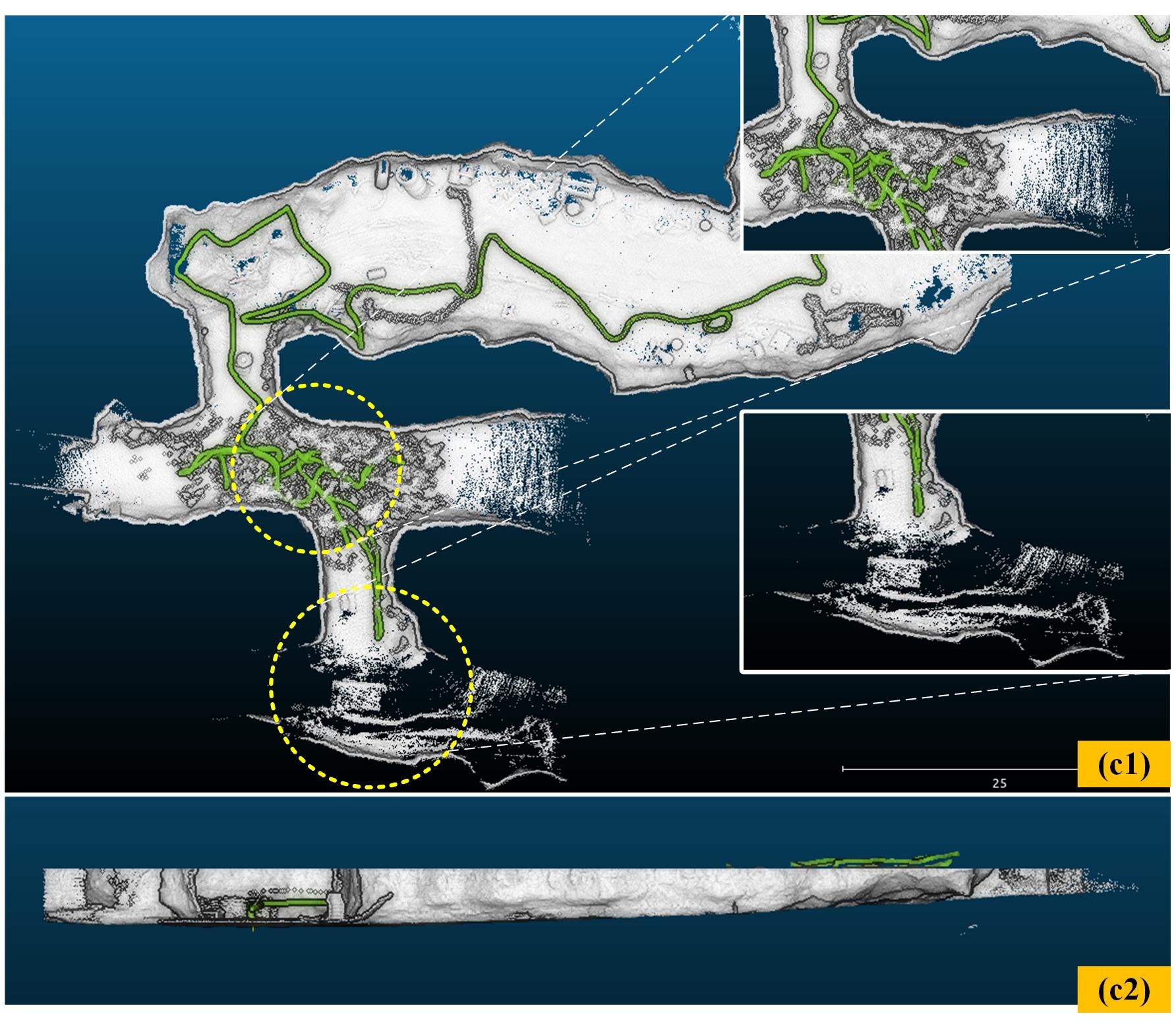}}
	\caption{Map comparison of LIO-SAM, LOAM, and Super Odometry in the dust data sequence. Upper image shows a top-down view and lower image shows the z-drift of the result map. We can see that Super Odometry is free of misalignment and gives a higher quality map.}
	\label{fig:dust-map}
	\vspace{-2mm}
\end{figure*}

\subsubsection{\textbf{Robustness Comparison in Geometrically Degraded Environments}}  

In this experiment, we ran the algorithms on the  \textbf{Long-Corridor} data sequence Fig.\ref{fig:real_scene}.(a). This is a geometrically degenerate scenario. It is common in man-made environments such as apartment buildings, hospitals, and factories. The lack of geometric features and repeated structures may cause issues for LiDAR-based algorithms. 

Fig.\ref{fig:long-corridor-map} shows the resulting map from LOAM, LIO-SAM, and Super Odometry in this data sequence. On one hand, we observe misalignment and significant drift as shown in the yellow circle in the LIO-SAM and LOAM maps. Super Odometry, on the other hand, outperforms the previous two methods and gives an accurate map and odometry result. 
We also evaluate our method on similarly geometrically-degraded situations including \textbf{Long Corridor} and \textbf{Constrained-Environment} data sequence mentioned in dataset section. The ATE of each method can be found in Table \ref{tab:accuracy}.
It can be seen that Super Odometry outperforms the other methods with the lowest ATE of 0.055m.  
For more details, we refer readers to the accompanying video.   
\subsubsection{\textbf{Robustness Comparison in Both Visually and Geometrically Degraded Environments}} 
Here we ran the algorithms on the \textbf{Dust} dataset. This experiment shows the algorithms' abilities to overcome airborne obscurants. The dust introduces challenges to visual odometry by reducing the number of useful features in the image. It is also a problem for LiDAR odometry because it blocks the sensor field-of-view with dust particles and thus we obtain fewer geometric features. Fig.\ref{fig:real_scene}(d) shows the large amount of dust generated from UAV's thrust in this experiment. Both vision-based methods (VINS-Mono and Depth-enhanced VINS) failed in this experiment as the camera provides limited information for state estimation
in the dust.

From Fig.\ref{fig:dust-map}, we can see that both LOAM and LIO-SAM fail to create an accurate map of the scene. In Fig.\ref{fig:dust-map}(a),
we can observe that the map generated by LOAM has significant misalignment as shown in the yellow circles. This is mainly because LOAM relies heavily on LiDAR
measurements and the IMU is only used to initialize the LiDAR scan matching. In the case of heavy dust, scan matching is likely to fail due to noisy returns. LIO-SAM performs better than LOAM, benefiting from its tightly coupled LiDAR-IMU architecture and reduced reliance on LiDAR measurements, although we can still observe map misalignment in the yellow circles in Fig.\ref{fig:dust-map}(b). Super Odometry outperforms both LOAM and LIO-SAM in this experiment. It gives an accurate and well-aligned map as shown in Fig.\ref{fig:dust-map}(c). The main reason for the superior result is that Super Odometry adopts a feature extraction method based on principal component analysis and evaluates the quality of each feature. Therefore, if a feature has low quality (see Eq.(\ref{eq:w1},\ref{eq:w2}) ) it will be regarded as noise, and not be used in state estimation.


\begin{table*}[htbp]
	\begin{center}
		\caption{Accuracy Evaluation of LOAM, LIO-SAM, VINS, VINS-Depth and Ours Methods Operating on Challenging Environments}
		\label{tab:ATE Trans.}
		\centering
		\setlength{\tabcolsep}{1mm}{
		\begin{tabular}{cccccc|ccccccccc}
			\toprule
			&\multicolumn{5}{c}{ATE(in m) Transl. MAX} &\multicolumn{5}{c}{ATE(in m) Transl. RMSE} \\ \midrule
			Sequences                & LOAM & LIO-SAM & VINS &VINS-Depth & Ours   & LOAM & LIO-SAM & VINS &VINS-Depth & Ours  \\ \midrule
			Constrained environments  & 0.779 & 1.914 & 0.907  & 0.972. & \bf0.609  & 0.319 & 0.650 & 0.470 & 0.490 &\bf 0.259  \\
			Long corridor             & 9.44 & 6.52  & 9.02 & 6.15 &\bf0.35 & 4.15 &0.85 & 5.83 & 2.58  &\bf0.055 \\
			White Wall                & 1.013 & 1.159 & 2.801 & 1.733 & \bf0.482   & 0.457 & 0.217 & 1.184 & 0.786 & \bf 0.156 \\
			Dark Room                 & 0.528 & 0.839 & 0.948 & 0.924 & \bf0.354 & 0.263 & 0.246 &0.527 & 0.429 & \bf0.174 \\
			\midrule
		\end{tabular}}
		\label{tab:accuracy}
		\vspace{-4mm}
	\end{center}
\end{table*}




\subsection{Real-time Performance Evaluation}
In this section, we show the real-time performance of Super Odometry and its ability to run efficiently on lightweight onboard systems. We benchmarked
the algorithms on an 8-core Intel Core i7-4790K CPU. First, we evaluate the impact of using our dynamic octree instead of the traditional KD-tree. Then, we evaluate the processing time of each submodule.

\subsubsection{\textbf{Dynamic Octree Performance}}
In order to discuss the performance of our dynamic octree versus traditional KD-tree, we compared the insertion and query time of dynamic octree and KD-tree in outdoor environments.
\begin{figure}[htbp]
	\centering
	\includegraphics[width=1.0\columnwidth]{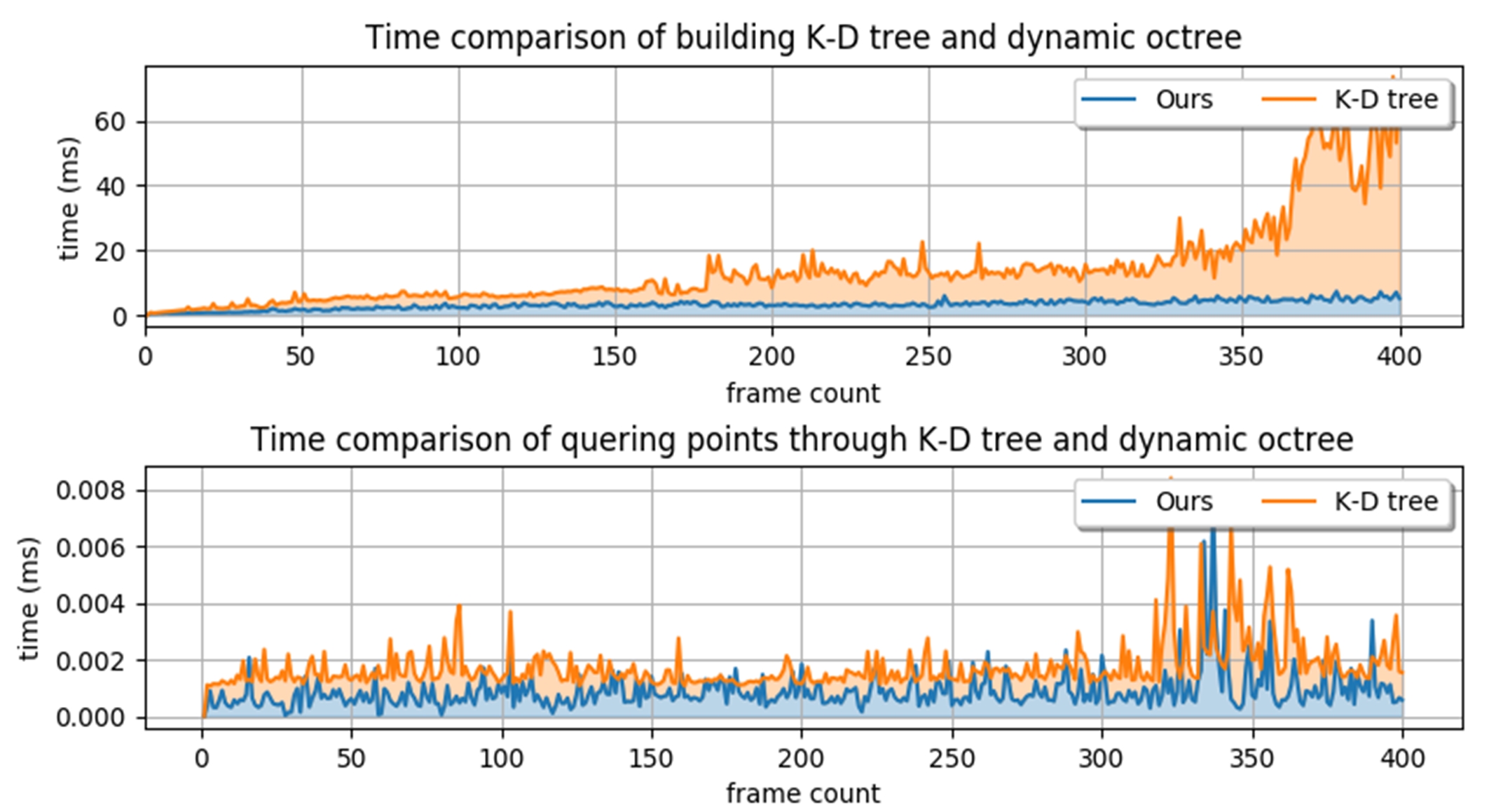}
	\caption{Runtime performance of dynamic octree vs. KD-tree in outdoor environment.}
	\label{fig:tree_outdoor}
\end{figure}
\vspace{-1mm}

As we can see in Fig.\ref{fig:tree_outdoor}, although query time remains similar for dynamic octree and KD-tree, tree-building times are significantly different, where KD-tree's runtime grows exponentially with the number of frames and dynamic octree's runtime remains almost constant. This is because dynamic
octree can avoid rebuilding the entire tree and only update the relevant sub-tree through hash map lookups.

\subsubsection{\textbf{Submodule Runtime}}
We evaluated each submodule processing time of Super Odometry and compared our LiDAR-inertial Odometry with popular LIO-SAM algorithm in various datasets shown as in  Table \ref{tab::runtime}. Overall, Super Odometry can achieve very high-frequency state estimation since all the submodules run in parallel. Instead of processing data in a sequential manner, it is important to note that super odometry achieves sensor fusion asynchronously. The optimizer of each submodule will cache the constraints and process them in small batches at its own frequency. Therefore, the total running time will be decided by the maximum submodule processing time, which significantly improves real-time performance. 

IMU Odometry can directly output 1000 Hz state estimation. The LiDAR process module of Super Odometry takes a smaller amount of time than LIO-SAM. The difference is more significant in the \textbf{Dust} dataset, where LIO-SAM 
averages 102 ms while Super Odometry averages 21 ms. This is because in dusty environments, LiDAR scan matching is slow to converge,
and Super Odometry can reject some unreliable constraints and better leverage the faster insertion and query of the dynamic octree.


\begin{table}[]
	\centering
	\caption{The average running time of Super Odometry in various dataset on desktop PC (with Intel Core i7-4790K CPU (ms))}
	\label{tab::runtime}
	\resizebox{0.9\columnwidth}{!}{
		\begin{tabular}{@{}cccccc@{}}
			\toprule
			Dataset   & LIO  & VIO  & IMUODOM   & Ours  &LIO-SAM  \\ \midrule
			Dust          & 21  & bypass      & 0.8   &21    & 102          \\
			Alpha Course    & 14  & bypass   & 0.9 & 14   & 28         \\
			Dark Room     & 13  & bypass       & 0.8   & 13   &23         \\
			Long Corridor  & 14  & 28    & 0.8     & 28   & 24         \\ \bottomrule
		\end{tabular}
	}
	\vspace{-3mm}
\end{table}

\section{Conclusion} 
\label{sec:conclusion}
Robust state estimation in perceptually-degraded environments is very challenging due to the lack of reliable measurements.
To solve this problem, we propose Super Odometry, an IMU-centric data processing pipeline, combining the advantages of a loosely coupled method with tightly coupled methods. To improve real-time performance, we propose to use the dynamic octree that significantly improves real-time performance.
The proposed method has been tested in both visually and geometrically degraded environments. 
The results show that our framework is robust to individual sensor failures and enables to achieve high-precision and resilient motion estimation in challenging real-world scenarios.


%

\vspace{-1mm}

\section*{ACKNOWLEDGMENT}
This research was sponsored by DARPA (\#HR00111820044). Content is not endorsed by and does not necessarily reflect the position or policy of the government or our sponsors.

\bibliographystyle{IEEEtran}
\bibliography{mybibfile}

\begin{thebibliography}{10}
\providecommand{\url}[1]{#1}
\csname url@samestyle\endcsname
\providecommand{\newblock}{\relax}
\providecommand{\bibinfo}[2]{#2}
\providecommand{\BIBentrySTDinterwordspacing}{\spaceskip=0pt\relax}
\providecommand{\BIBentryALTinterwordstretchfactor}{4}
\providecommand{\BIBentryALTinterwordspacing}{\spaceskip=\fontdimen2\font plus
\BIBentryALTinterwordstretchfactor\fontdimen3\font minus
  \fontdimen4\font\relax}
\providecommand{\BIBforeignlanguage}[2]{{%
\expandafter\ifx\csname l@#1\endcsname\relax
\typeout{** WARNING: IEEEtran.bst: No hyphenation pattern has been}%
\typeout{** loaded for the language `#1'. Using the pattern for}%
\typeout{** the default language instead.}%
\else
\language=\csname l@#1\endcsname
\fi
#2}}
\providecommand{\BIBdecl}{\relax}
\BIBdecl

\bibitem{loam}
J.~Zhang and S.~Singh, ``Loam: Lidar odometry and mapping in real-time.'' in
  \emph{Robotics: Science and Systems}, vol.~2, no.~9, 2014.

\bibitem{shan2020liosam}
T.~Shan, B.~Englot, D.~Meyers, W.~Wang, C.~Ratti, and D.~Rus, ``Lio-sam:
  Tightly-coupled lidar inertial odometry via smoothing and mapping,'' 2020.

\bibitem{Zhao_2019}
\BIBentryALTinterwordspacing
S.~Zhao, Z.~Fang, H.~Li, and S.~Scherer, ``A robust laser-inertial odometry and
  mapping method for large-scale highway environments,'' \emph{2019 IEEE/RSJ
  International Conference on Intelligent Robots and Systems (IROS)}, Nov 2019.
  [Online]. Available: \url{http://dx.doi.org/10.1109/IROS40897.2019.8967880}
\BIBentrySTDinterwordspacing

\bibitem{VINS}
T.~Qin, P.~Li, and S.~Shen, ``Vins-mono: A robust and versatile monocular
  visual-inertial state estimator,'' \emph{IEEE Transactions on Robotics},
  vol.~34, no.~4, pp. 1004--1020, 2018.

\bibitem{zhao2020tptio}
S.~Zhao, P.~Wang, H.~Zhang, Z.~Fang, and S.~Scherer, ``Tp-tio: A robust
  thermal-inertial odometry with deep thermalpoint,'' 2020.

\bibitem{graeter2018limo}
J.~Graeter, A.~Wilczynski, and M.~Lauer, ``Limo: Lidar-monocular visual
  odometry,'' in \emph{2018 IEEE/RSJ International Conference on Intelligent
  Robots and Systems (IROS)}.\hskip 1em plus 0.5em minus 0.4em\relax IEEE,
  2018, pp. 7872--7879.

\bibitem{ebadi2020lamp}
K.~Ebadi, Y.~Chang, M.~Palieri, A.~Stephens, A.~Hatteland, E.~Heiden,
  A.~Thakur, N.~Funabiki, B.~Morrell, S.~Wood, L.~Carlone, and A.~akbar
  Agha-mohammadi, ``Lamp: Large-scale autonomous mapping and positioning for
  exploration of perceptually-degraded subterranean environments,'' 2020.

\bibitem{Palieri_2021}
\BIBentryALTinterwordspacing
M.~Palieri, B.~Morrell, A.~Thakur, K.~Ebadi, J.~Nash, A.~Chatterjee,
  C.~Kanellakis, L.~Carlone, C.~Guaragnella, and A.-a. Agha-mohammadi, ``Locus:
  A multi-sensor lidar-centric solution for high-precision odometry and 3d
  mapping in real-time,'' \emph{IEEE Robotics and Automation Letters}, vol.~6,
  no.~2, p. 421–428, Apr 2021. [Online]. Available:
  \url{http://dx.doi.org/10.1109/LRA.2020.3044864}
\BIBentrySTDinterwordspacing

\bibitem{zhang2018laser}
J.~Zhang and S.~Singh, ``Laser--visual--inertial odometry and mapping with high
  robustness and low drift,'' \emph{Journal of Field Robotics}, vol.~35, no.~8,
  pp. 1242--1264, 2018.

\bibitem{multi_modal}
S.~{Khattak}, H.~{Nguyen}, F.~{Mascarich}, T.~{Dang}, and K.~{Alexis},
  ``Complementary multi–modal sensor fusion for resilient robot pose
  estimation in subterranean environments,'' in \emph{2020 International
  Conference on Unmanned Aircraft Systems (ICUAS)}, 2020, pp. 1024--1029.

\bibitem{camurri2020pronto}
M.~Camurri, M.~Ramezani, S.~Nobili, and M.~Fallon, ``Pronto: a multi-sensor
  state estimator for legged robots in real world scenarios,'' \emph{Frontiers
  in Robotics and AI}, vol.~7, p.~68, 2020.

\bibitem{Reijgwart_2020}
\BIBentryALTinterwordspacing
V.~Reijgwart, A.~Millane, H.~Oleynikova, R.~Siegwart, C.~Cadena, and J.~Nieto,
  ``Voxgraph: Globally consistent, volumetric mapping using signed distance
  function submaps,'' \emph{IEEE Robotics and Automation Letters}, vol.~5,
  no.~1, p. 227–234, Jan 2020. [Online]. Available:
  \url{http://dx.doi.org/10.1109/LRA.2019.2953859}
\BIBentrySTDinterwordspacing

\bibitem{shao2019stereo}
W.~Shao, S.~Vijayarangan, C.~Li, and G.~Kantor, ``Stereo visual inertial lidar
  simultaneous localization and mapping,'' 2019.

\bibitem{zuo2020licfusion}
X.~Zuo, Y.~Yang, P.~Geneva, J.~Lv, Y.~Liu, G.~Huang, and M.~Pollefeys,
  ``Lic-fusion 2.0: Lidar-inertial-camera odometry with sliding-window
  plane-feature tracking,'' 2020.

\bibitem{msckf}
A.~I. Mourikis and S.~I. Roumeliotis, ``A multi-state constraint kalman filter
  for vision-aided inertial navigation,'' in \emph{Proceedings 2007 IEEE
  International Conference on Robotics and Automation}, 2007, pp. 3565--3572.

\bibitem{Wisth_2021}
\BIBentryALTinterwordspacing
D.~Wisth, M.~Camurri, S.~Das, and M.~Fallon, ``Unified multi-modal landmark
  tracking for tightly coupled lidar-visual-inertial odometry,'' \emph{IEEE
  Robotics and Automation Letters}, vol.~6, no.~2, p. 1004–1011, Apr 2021.
  [Online]. Available: \url{http://dx.doi.org/10.1109/LRA.2021.3056380}
\BIBentrySTDinterwordspacing

\bibitem{forster2015imu}
C.~Forster, L.~Carlone, F.~Dellaert, and D.~Scaramuzza, ``Imu preintegration on
  manifold for efficient visual-inertial maximum-a-posteriori
  estimation.''\hskip 1em plus 0.5em minus 0.4em\relax Georgia Institute of
  Technology, 2015.

\bibitem{hackel2016fast}
T.~Hackel, J.~D. Wegner, and K.~Schindler, ``Fast semantic segmentation of 3d
  point clouds with strongly varying density,'' \emph{ISPRS annals of the
  photogrammetry, remote sensing and spatial information sciences}, vol.~3, pp.
  177--184, 2016.

\bibitem{behley2015icra}
J.~Behley, V.~Steinhage, and A.~B. Cremers, ``{Efficient Radius Neighbor Seach
  in Three-dimensional Point Clouds},'' in \emph{Proc. of the IEEE
  International Conference on Robotics and Automation (ICRA)}, 2015.

\bibitem{vinsicra}
P.~W. Zheng~Fang, Shibo~Zhao, ``{Vanishing Point Aided LiDAR-Visual-Inertial
  Estimator},'' in \emph{Proc. of the IEEE International Conference on Robotics
  and Automation (ICRA)}, 2021.

\end{thebibliography}

\end{document}